\documentclass[10pt,journal,compsoc]{IEEEtran}

\ifCLASSOPTIONcompsoc
  \usepackage[nocompress]{cite}
\else
  \usepackage{cite}
\fi

\usepackage[pdftex]{graphicx}
\graphicspath{res}
\DeclareGraphicsExtensions{.pdf,.jpg,.png}

\usepackage{algorithm}
\floatstyle{plaintop} \restylefloat{algorithm}
\usepackage{algorithmic}
\usepackage{amsmath}
\usepackage{array}
\usepackage{booktabs}
\usepackage{subcaption}
\usepackage{subfloat}
\usepackage{url}

\newcolumntype{M}[1]{>{\centering\arraybackslash}m{#1}}

\begin{document}


\title{Point Proposal Network: Accelerating Point Source Detection Through Deep Learning}

\author{Duncan~Tilley,~
        Christopher~W.~Cleghorn,~\textit{Member},~\textit{IEEE},~
        Kshitij~Thorat,~
        and~Roger~Deane
\IEEEcompsocitemizethanks{
    \IEEEcompsocthanksitem D. Tilley and C. W. Cleghorn are with the Department of Computer Science, University of Pretoria, Pretoria, South Africa.
    \protect\\(Contact: see http://tilleyd.com; ccleghorn@cs.up.ac.za)
    \IEEEcompsocthanksitem K. Thorat and R. Deane are with the Department of Physics, University of Pretoria, Pretoria, South Africa.
    \protect\\(Contact: thorat.k@gmail.com; roger.deane@up.ac.za)}%
}


\markboth{}
{Tilley \MakeLowercase{\textit{et al.}}: PPN: Accelerating Point Source Detection Through Deep Learning}

\IEEEtitleabstractindextext{%
\begin{abstract}
    Point source detection techniques are used to identify and localise point sources in radio astronomical surveys. With the development of the Square Kilometre Array (SKA) telescope, survey images will see a massive increase in size from Gigapixels to Terapixels. Point source detection has already proven to be a challenge in recent surveys performed by SKA pathfinder telescopes. This paper proposes the Point Proposal Network (PPN): a point source detector that utilises deep convolutional neural networks for fast source detection. Results measured on simulated MeerKAT images show that, although less precise when compared to leading alternative approaches, PPN performs source detection faster and is able to scale to large images, unlike the alternative approaches.
\end{abstract}

}

\maketitle

\IEEEdisplaynontitleabstractindextext

\IEEEraisesectionheading{\section{Introduction}\label{sec:introduction}}

\IEEEPARstart{T}{he} long wavelengths observed in radio astronomy allow wide fields of view to be imaged with modern interferometers, while being sensitive to important radio source populations, including radio galaxies, supernovae, and a suite of explosive phenomena across the Universe. The Square Kilometre Array (SKA) telescope is a proposed radio interferometer that, when completed, will be able to conduct surveys with an unparallaled combination of sensitivity, angular resolution and imaging fidelity \cite{Norris2013}.

The SKA pathfinders such as the MeerKAT in South Africa and the ASKAP in Australia are already conducting both wider and deeper surveys than before, providing a steady build-up to the arrival of the SKA. In addition, wide-field Very Long Baseline Interferometry (VLBI) is a burgeoning field which routinely produces Terapixel images within a single field of view.

It is a necessary step in the scientific pipeline to extract properties of observable objects, or \textit{sources} from a survey image. At the very least, the position of a candidate source must be known to perform any further processing. Detecting or classifying sources are in some cases done by hand, as is the noble goal of the Radio Galaxy Zoo \cite{Banfield2015}. For the sake of time, however, automated techniques are used which usually provide the minimal information necessary to describe point sources.

As radio telescope survey speeds and angular resolutions increase, the size of surveyed images increases. For this reason, recent years have seen the introduction of many new source detection techniques which attempt to improve the reliability \cite{Hales2012}\cite{Mohan2015}\cite{Sadr2018}, or in few cases, the speed \cite{Carbone2018} of previous techniques. Many recent surveys have also analysed the performance of existing source detection techniques when applied to modern telescopes \cite{Hancock2012}\cite{Huynh2012}\cite{Hopkins2015}. These surveys emphasise the growing need for novel detection methods.

This paper limits its investigation to \textit{point} sources rather than \textit{extended} sources, the former having an extent significantly smaller than the instrument's point spread function full width at half maximum (FWHM). This limitation is justified by the fact that point sources dominate detections in current surveys from the MeerKAT and ASKAP.

Most point source detectors roughly follow the same skeleton consisting of 3 steps \cite{Hancock2012}. The following is a brief description of each step:
\begin{enumerate}
    \item \textit{Background estimation}, whereby the background noise of an image is estimated and subtracted from the original image in order to obtain a clearer image \cite{Bertin1996}.
    \item \textit{Source detection}, which usually involves performing the computationally expensive flood-fill algorithm to detect contiguous islands of pixels that lie above a certain threshold. These islands are then identified as candidate sources.
    \item \textit{Source characterisation}, in which various characteristics of each source are calculated. The simplest of these is the sky coordinates (right ascension, RA and declination, DEC) of the point source's centre.
\end{enumerate}

Hancock et~al. \cite{Hancock2012} add the final step of cataloguing sources which will be considered as part of source characterisation here, as few detectors perform any notable post-processing. As mentioned, the flood-filling approach utilised by point source detectors is expensive, which doesn't scale well to larger images.

This paper proposes a new point source detection technique that utilises modern advances in object detection. The technique is called the \textit{Point Proposal Network} (PPN), which uses a deep convolutional neural network (CNN) \cite{LeCun1998}\cite{Krizhevsky2012}\cite{Goodfellow2016} to perform all three steps lined out above in a single evaluation of the CNN. The technique therefore eliminates the need of a flood filling iteration and aims to provide a faster and more scalable point source detection method for use in current and future large surveys.

The remainder of this paper is laid out as follows. Section \ref{sec:related} discusses related works in point source and object detection. Section \ref{sec:method} describes the PPN approach in full. Section \ref{sec:results} discusses the experiments that were performed and illustrates the performance of PPN in comparison to DeepSource \cite{Sadr2018}, both in terms of accuracy and speed. Finally, Section \ref{sec:conclusion} provides concluding remarks.

\section{Related Work}
\label{sec:related}

\subsection{Point Source Detectors}
\label{sec:related:psd}

SExtractor \cite{Bertin1996} is an early point source detector that first introduced the skeleton discussed in the introduction. This point source detector influenced virtually all later point source detectors, with many of them introducing only small variants to the different steps \cite{Mohan2015}\cite{Hales2012}\cite{Carbone2018}.

DeepSource is a point source detector that was recently proposed by Sadr et~al. \cite{Sadr2018}. To the authors' knowledge, it is the first point source detector that involves neural networks within the detection pipeline. DeepSource uses a CNN to perform background noise estimation by directly transforming the image to a cleaner version. This has improved the accuracy of point source detections in the presence of noise; however, the source detection is still performed using a variant of flood filling and suffers from the same poor scaling as earlier techniques.

\subsection{Object Detectors}
\label{sec:related:object}

R-CNN is one of the first object detectors that utilised a deep CNN \cite{Girshick2014} and led to the development of many other modern object detectors. A notable variant of this technique is Faster R-CNN \cite{Ren2015} which uses a CNN to perform localisation by proposing regions of possible objects. This CNN is aptly named the Region Proposal Network (RPN).

The output of the RPN is then used to generate subsets of the input image, which are fed through additional layers to perform classification. The use of an RPN greatly improves on the speed of R-CNN, but multiple evaluations of the later layers are still required for each proposed area. Other methods have proposed ways to perform both localisation and classification in a single evaluation of a CNN, such as YOLO and SSD \cite{Redmon2016}\cite{Redmon2017}\cite{Liu2016}.

The approaches used by these techniques allow for faster object detection. This makes them strong candidates for situations where high-speed detection of objects is required, such as in videos or possibly very large images. These techniques are what led to the point source detector proposed in this paper.

\section{Point Proposal Network}
\label{sec:method}

In this section, the details of PPN are discussed\footnote{An implementation of PPN including data simulation can be found at \url{https://github.com/tilleyd/point-proposal-net}}. Section \ref{sec:method:ppn} first fully describes the technique, with Section \ref{sec:method:param} showing the steps taken to choose the values for hyper-parameters and develop the full network architecture.

\subsection{Proposed Method}\label{sec:method:ppn}

The PPN approach described here considers point source detection as a regression problem. The network receives a single image as input and directly provides proposed positions of point sources. The technique therefore does not follow the traditional point source detection steps and provides a faster way of performing point source detection.

The architecture of PPN is based on the Faster-RCNN Region Proposal Network (RPN) \cite{Ren2015}. The RPN is specifically designed for only proposing the regions of objects and does not include classification information in the direct output, making it suitable for this problem. In addition, the RPN architecture and loss function are fairly loosely defined, allowing modification to perform point proposals instead of region proposals.

\subsubsection{Architecture}

Similar to the RPN, PPN performs regression based on \textit{anchors} placed across the input image. However, in the case of PPN each anchor only outputs a single offset to a possible point source's centre and doesn't include any bounding box information. To this end, the anchors will instead be referred to as the offset \textit{origins}.

PPN is separated into two logical parts: the \textit{base layers} and the \textit{proposal layers}. The base layers can have an arbitrary architecture and are explored later in this section. The shape of the base layers' output feature map defines the number of origins. If the output feature map is of shape $(m, n, k)$, then there will be $m$ rows and $n$ columns of origins. The number of output channels, $k$, does not affect the shape of the final output.

\begin{figure}
\begin{center}
    \centerline{\includegraphics[scale=0.5]{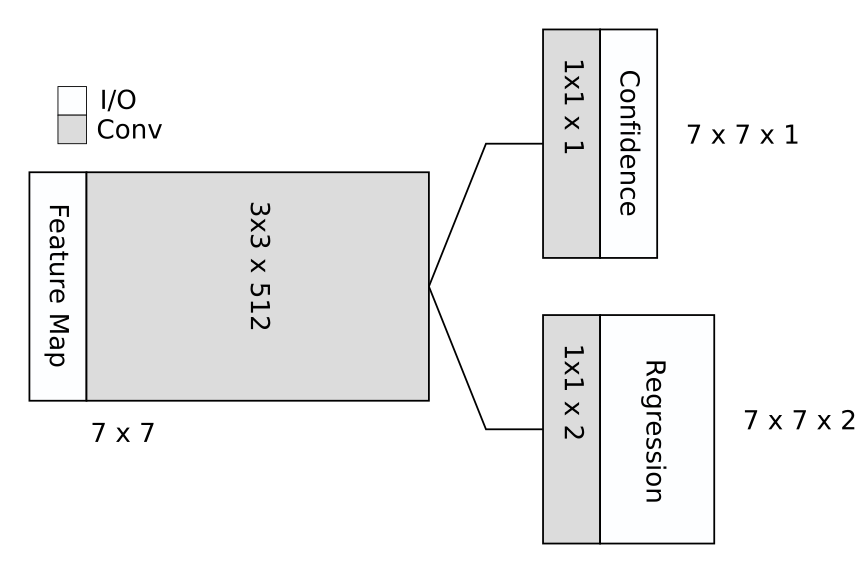}}
    \caption{An illustration of the 3 convolutional layers responsible for producing point proposals from the base layers' feature map. The layers use the features extracted by the base layers and creates separate confidence and regression matrices. These layers are zero-padded with no activation functions. Only the confidence output is passed through a Sigmoid activation function to produce an output between 0 and 1.}
    \label{fig:proposal}
\end{center}
\vskip -0.2in
\end{figure}

The proposal layers are a number of convolutional layers that use the features extracted by the base layers to perform regression. The final output of the proposal layers is an $(m \times n \times 2)$ regression matrix, and an $(m \times n \times 1)$ confidence matrix. The architecture of these layers is shown in Fig.~\ref{fig:proposal}.

The last dimension of the regression matrix contains tuples $(\Delta x, \Delta y)$, which are the offsets from origin $(m, n)$'s position to a possible point source. The offsets are normalised such that an offset of 1 or -1 will line up with the neighbouring origin. The last dimension of the confidence matrix contains a probability score for origin $(m, n)$, which indicates the probability that there is in fact a point at the position pointed to by $(\Delta x, \Delta y)$.

\begin{figure}
\begin{center}
    \centerline{\includegraphics[scale=0.5]{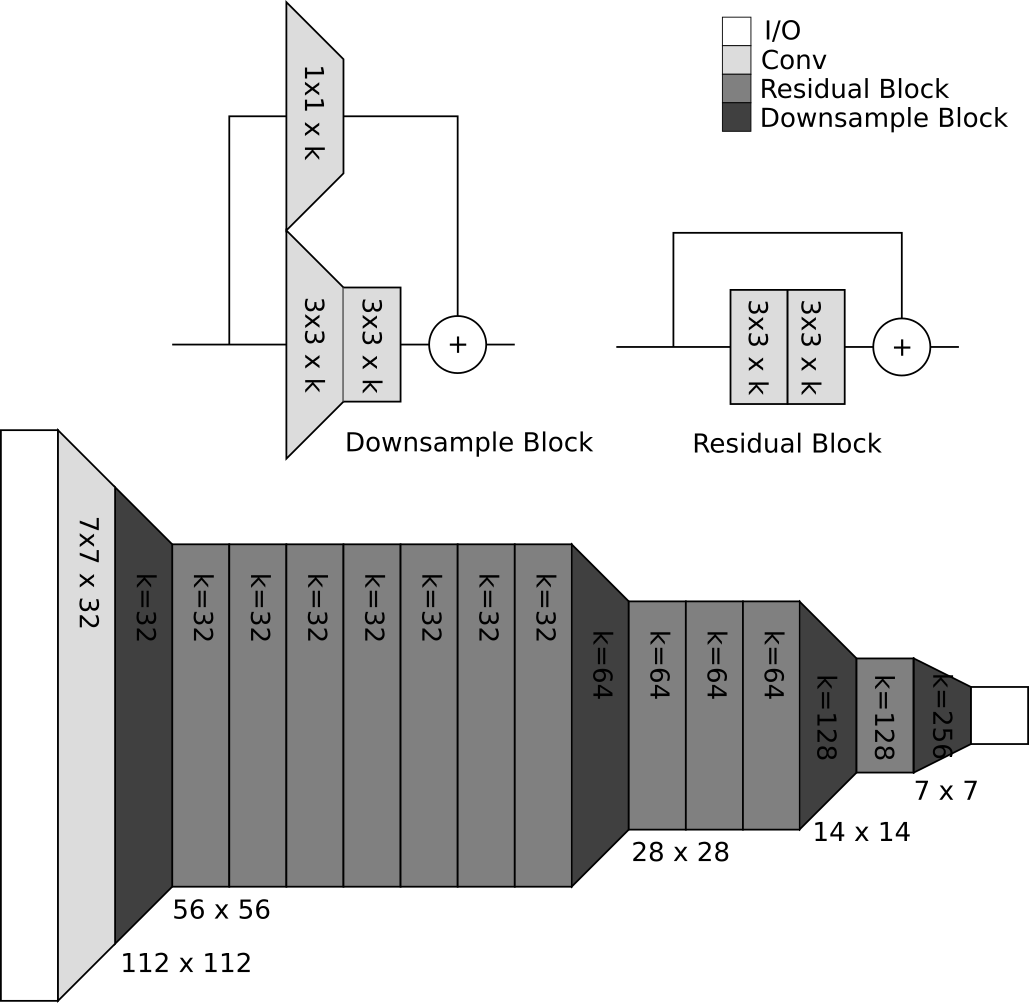}}
    \caption{The 31-layer ResNet base layer architecture. Each convolutional layer is zero-padded and followed by batch normalisation, before being passed through a ReLU activation function. The slanted convolutional layers use a stride of 2 in order to down-sample the input feature map.}
    \label{fig:resnet}
\end{center}
\vskip -0.2in
\end{figure}

The base layers of PPN are crucial to effectively extract features from images. Experiments with the base layers (see Section~\ref{sec:method:param:base}) showed the best performing architecture among those evaluated is a 31-layer ResNet \cite{He2016} architecture, as illustrated in Fig.~\ref{fig:resnet}. More residual blocks were added to the earlier layers in order to minimise the amount of information loss when down-sampling the feature maps. This has shown to perform slightly better than giving an equal number of blocks to each section, possibly due to the small size of point sources being highly susceptible to information loss.

\subsubsection{Training}

PPN uses a supervised learning approach to minimise the loss function. Since the output of PPN is two separate matrices, the loss function consists of two parts. The first part uses focal loss \cite{Lin2017} for the confidence output \eqref{eq:conf_loss} and the second uses the mean-squared error for the regression output \eqref{eq:reg_loss}.

The use of focal loss allows more control over the landscape of the loss function by exponentially increasing the loss contribution of ``difficult'' samples and decreasing that of ``easy'' samples. This is controlled by the $\gamma$ hyper-parameter. Additionally, an $\alpha$ weighting factor is used for positive samples. The choice of the focal loss hyper-parameters $\alpha$ and $\gamma$ is explored in Section~\ref{sec:method:param:focal}.

\begin{align}
    E_c &= -\sum_m\sum_n b_{mn} \cdot \begin{cases}
        \alpha(1-c_{mn})^\gamma\log(c_{mn}) & \mathrm{if\ } \hat{c}_{mn} = 1\\
        c_{mn}^\gamma\log(1-c_{mn}) & \mathrm{otherwise}
    \end{cases} \label{eq:conf_loss} \\
    E_r &= \frac{1}{m\cdot n} \sum_m\sum_n b^*_{mn} \cdot ||\hat{\boldsymbol{r}}_{mn} - \boldsymbol{r}_{mn}||_2^2 \label{eq:reg_loss}
\end{align}

The confidence ground truth, $\hat{c}$, is 1 for origins which are within a radius $r_{\mathrm{near}}$ from a point source, and 0 if there is no point source close enough. The regression ground truth, $\hat{\boldsymbol{r}}$, is the normalised offset from the origin to the nearest point source and is only calculated if $\hat{c} = 1$.

$b$ and $b^*$ are flag values that are either 1 or 0. This allows control over which origins should contribute to each part of the loss function. $b$ is 1 only for origins with $\hat{c} = 1$, or where the nearest point source is further than $r_\mathrm{far}$. $b^*$ is 1 only for origins with $\hat{c} = 1$.

The flags $b$ and $b^*$ allow only origins that are sufficiently close to a point source to contribute to the regression loss, and only origins that are either sufficiently close or sufficiently far to contribute to the confidence loss. Origins that are in the range $(r_\mathrm{near}, r_\mathrm{far})$ from the nearest point source will not contribute to the loss function.

The confidence and regression losses are combined as in Eq.~\ref{eq:ppn_loss}. Following the technique used by the RPN, $N_c$ and $N_r$ are balancing coefficients introduced to ensure that both parts have an equal weighting. The Faster R-CNN authors \cite{Ren2015} recommend to simply use $N_r = N_c = \frac{1}{\mathrm{batch\ size}}$. These recommended values do not perform any balancing, but will instead result in the loss values being averaged across the batch.

\begin{equation}\label{eq:ppn_loss}
    E = N_cE_c + N_rE_r
\end{equation}

For this paper, all training was done using the Adam optimiser \cite{Kingma2014} with a drop rate \cite{Srivastava2014} of 20\% in all layers. A training set of 8192 simulated image patches (see Sections \ref{sec:results:images}) are used with a batch size of 128. Additionally, validation and testing sets of 512 image patches each are used to validate the model. During training, the model with the lowest validation loss value is saved and recalled at the end before testing.

\subsubsection{Inferencing}
\label{sec:method:ppn:inferencing}

The PPN input has a fixed shape, so all images given to the network must have the same shape. It is impractical to scale any image to the required size as this will cause a lot of information to be lost. To allow the use on different sized images, the input image is segmented into smaller patches, which may optionally overlap. The patches are then individually passed through the network to obtain the output proposals for each patch.

The regression output is in the form of normalised offsets, which must first be denormalised back into pixel coordinates. Next, the positions of each individual origin must be added to the regression offsets, which results in positions within the patch. To obtain positions within the original image, the patch's origin must also be added to all the positions. These are all matrix additions, which can be efficiently computed on a GPU and are incorporated directly into the PPN model.

A problem that arises with PPN is that there are a large amount of duplicate proposals made. Multiple origins may propose the same point, and duplicates may also arise as a result of patching (whereby points at the edges of overlapping patches are proposed in both patches). Duplicate removal is performed on the combined proposals for the entire image using a form of \textit{Non-Maximum Suppression} (NMS). In short, the algorithm takes the most confident point and removes all points within a radius $r_\mathrm{NMS}$ from that point. During this process, points with a confidence score below $c_{\mathrm{NMS}}$ are also removed. The procedure when applied to points $P$ is presented in the following algorithm.

\begin{algorithm}[H]
\begin{algorithmic}
\STATE {\bfseries Input:} $P, r_\mathrm{NMS}, c_\mathrm{NMS}$
\STATE $P\gets \{\boldsymbol{p} \in P : c(\boldsymbol{p}) \ge c_\mathrm{NMS}\}$
\STATE $P^* \gets \emptyset$
\WHILE {$P \neq \emptyset$}
    \STATE $\boldsymbol{p} \gets \mathrm{argmax}_{\boldsymbol{p}\in P}(c(\boldsymbol{p}))$
    \STATE $P^* \gets P^* \cup \{\boldsymbol{p}\}$
    \STATE $P \gets P \setminus \{\boldsymbol{r} \in P: ||\boldsymbol{r} - \boldsymbol{p}|| < r_\mathrm{NMS}\}$
\ENDWHILE
\STATE {\bfseries Return:} $P^*$
\end{algorithmic}
\end{algorithm}

\subsubsection{Performance Metrics}\label{sec:method:metrics}

In order to measure the performance of PPN, multiple metrics are taken into account. \textit{Precision} and \textit{recall} metrics are primarily used, defined in Eq.~\eqref{eq:prec_rec}. Precision is the ratio of the correct detections to the total number of detections made, measuring how reliable the results are. Recall is the ratio of correct detections to the total number of point sources in the image, measuring how complete the results are.

\begin{align}\label{eq:prec_rec}
    \mathrm{precision} = \frac{tp}{tp + fp} &&
    \mathrm{recall} = \frac{tp}{tp + fn}
\end{align}

A prediction will be considered a true positive when it is within a radius, $r_{tp}$, of the ground truth centre of a point source. Additionally, if either a ground truth or a prediction has been counted, it is excluded from further checks. This ensures that when two or more predictions are near a single source, only one will count as a true positive. Conversely, if two or more true sources surround a single prediction, only one match will be counted as a true positive.

Note that the value of $r_{tp}$ does not directly impact the performance of PPN, but rather determines how harsh the performance metrics are. The value of $r_{tp}$ is therefore fixed to 0.4 for the process of hyper-parameter tuning and for all recall measurements. In Section~\ref{sec:results}, the precision is evaluated at different values of $r_{tp}$, which shows how close the inferred sources are to the true centres.

In addition to precision and recall, the $F_1$ score is used as defined in Eq.~\eqref{eq:f1}. This provides a balanced combination of both precision and recall and is mostly used in this study to compare different configurations of hyper-parameters to choose the best balanced model. In practice however, it may be beneficial to consider either recall or precision to be more important.

\begin{equation}\label{eq:f1}
    F_1 = 2 \cdot \frac{\mathrm{precision} \cdot \mathrm{recall}}{\mathrm{precision} + \mathrm{recall}}
\end{equation}

\subsection{Hyper-Parameter Selection}\label{sec:method:param}

Hyper-parameter selection is important to maximise the performance of PPN. In order to choose good values for hyper-parameters, a manual search of different values was performed while attempting to maximise the precision and recall for images in the validation set. This section first discusses the selection of base layers in Section~\ref{sec:method:param:base}, then explores the values of all hyper-parameters used during both training and inferencing in Sections~\ref{sec:method:param:focal} and~\ref{sec:method:param:regular}.

\subsubsection{Base Architecture}\label{sec:method:param:base}

Three different architectures were initially evaluated during base layer selection. The first is a 13-layer VGGNet variant, which is similar to VGG16 \cite{Simonyan2014} but has the fully-connected layers dropped away. The second architecture is a 17-layer ResNet architecture. The third and final architecture is a custom 51-layer ``pyramid'' model, which builds on the premise that down-sampling should be minimised to prevent information loss. To this end, the architecture relies on the loss of size when no padding is used in convolutional layers to funnel down to the desired size.

\begin{figure}
\begin{center}
    \centerline{\includegraphics[scale=0.5]{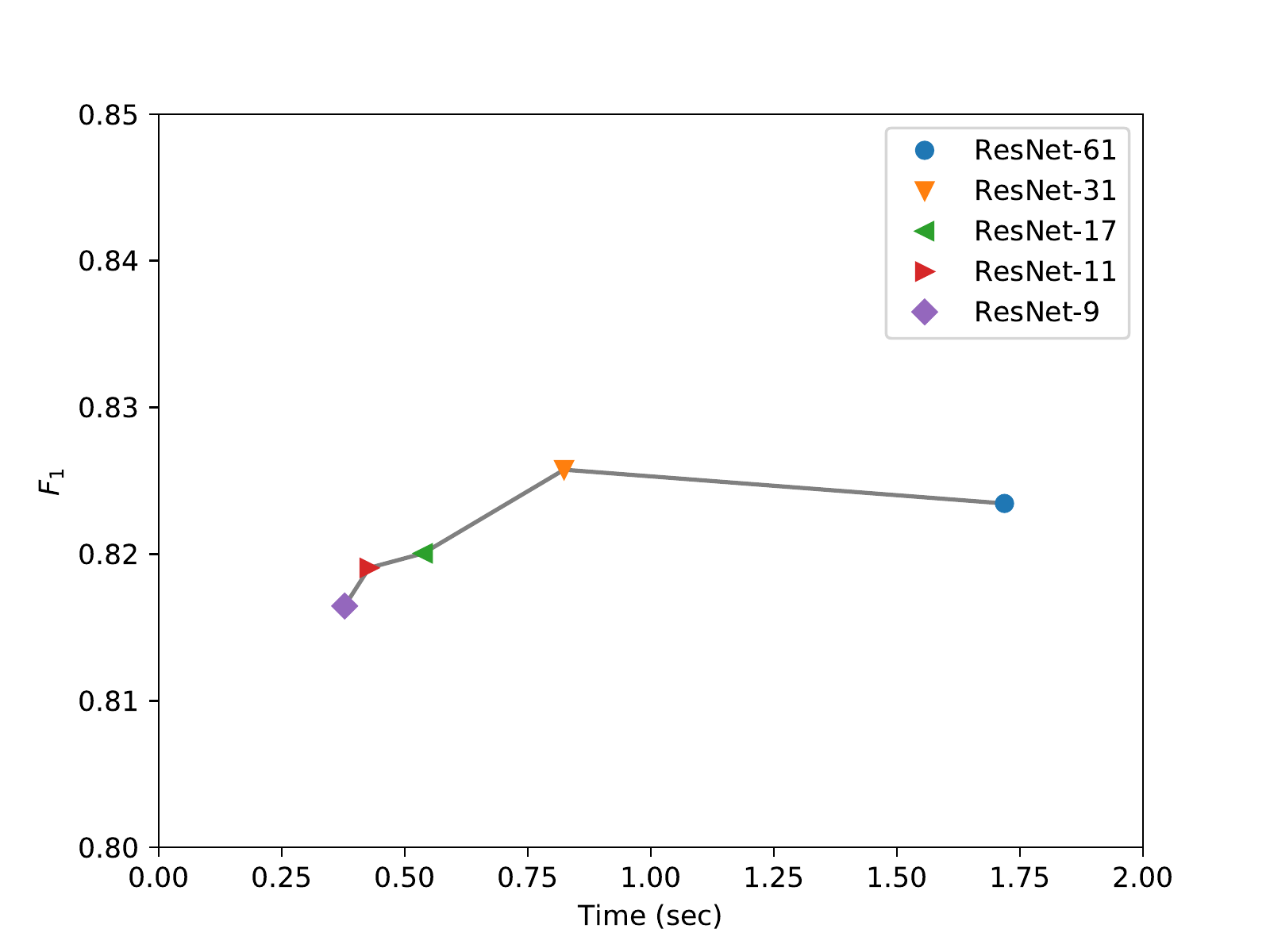}}
    \caption{The $F_1$ scores and inference speeds of PPN when using different sized ResNet architectures as base layers. Inference speeds were measured on $(4096 \mathrm{px} \times 4096 \mathrm{px})$ images.}
    \label{fig:size-speed}
\end{center}
\vskip -0.2in
\end{figure}

The results showed that the ResNet architecture is the best choice as it acquires a higher precision than VGGNet, though both are strong contenders for recall. The pyramid proved difficult to train and has poor performance in comparison to the other architectures. The ResNet architecture was explored further, with different sizes being tested. The results of testing a 9-layer up to a 61-layer ResNet are shown in Fig.~\ref{fig:size-speed}.

The graph shows the evaluation speed versus the $F_1$ score of the different architectures. These results show that there are very marginal improvements in the $F_1$ score up to the 31-layer ResNet, with it becoming worse with the 61-layer architecture. It may be beneficial to instead use the smaller architectures since even the 9-layer architecture is able to perform well, but ResNet-31 was chosen for the purposes of this paper as it boasts the highest $F_1$ score; however marginally.

\subsubsection{Focal Loss Parameters}\label{sec:method:param:focal}

The performance of focal loss \cite{Lin2017} as shown in Eq.~\eqref{eq:conf_loss} is sensitive to the values of $\alpha$ and $\gamma$, so different values were explored in combination. Following the recommendation of the authors, the biases of the final layer are initialized to $-\log\bigl((1-\pi)/\pi\bigr)$ when using focal loss, with $\pi = 0.01$. This forces the confidence matrix to provide a confidence roughly equal to $\pi$, increasing the loss contribution of easy negative samples during early training.

The measured performance of different models is shown in Table~\ref{tab:focal_params}. In the absence of focal loss ($\gamma = 0$), precision worsens as $\alpha$ increases with an improvement in recall. This is expected as the positive samples carry more weight, causing the network to learn to identify more difficult samples while also causing it to misidentify more noise. This pattern however disappears when $\gamma > 0$.

Increasing $\gamma$ tends to decrease recall, which is especially evident when comparing the results of $\gamma = 2$ to those of $\gamma = 0$. The precision also decreases slightly as $\gamma$ is increased, though the value increases as $\alpha$ increases. The shortcoming of focal loss in this case may be attributed to the ``difficult'' samples of this problem being particularly difficult; the faint sources are effectively indistinguishable from the background noise. The focal loss puts less focus on the easier samples while still being unable to correctly detect the difficult ones, leading to an overall worse recall.

As a result, the values of $\alpha = 0.5$ and $\gamma = 0$ are chosen as they maximise the $F_1$ score. With these values the loss function is of course equivalent to a normal weighted binary cross-entropy.

\begin{table*}
\caption{Precision, Recall and $F_1$ Scores of PPN for Different Focal Loss Parameters}
\label{tab:focal_params}
\begin{center}
\begin{small}
\begin{sc}
\begin{tabular}[c]{ccccc}
    \toprule
    $\gamma$ & $\alpha$ & Precision & Recall & $F_1$ \\
    \midrule
    0.0 & 0.5 & $\mathbf{0.947 \pm 0.040}$ & $0.733 \pm 0.040$ & $\mathbf{0.825 \pm 0.027}$ \\
    0.0 & 1.0 & $0.926 \pm 0.043$ & $0.742 \pm 0.033$ & $0.823 \pm 0.023$ \\
    0.0 & 2.0 & $0.876 \pm 0.064$ & $\mathbf{0.763 \pm 0.035}$ & $0.814 \pm 0.028$ \\
    1.0 & 0.5 & $0.722 \pm 0.166$ & $0.702 \pm 0.075$ & $0.707 \pm 0.116$ \\
    1.0 & 1.0 & $0.876 \pm 0.107$ & $0.737 \pm 0.036$ & $0.797 \pm 0.058$ \\
    1.0 & 2.0 & $0.887 \pm 0.092$ & $0.739 \pm 0.035$ & $0.804 \pm 0.052$ \\
    2.0 & 0.5 & $0.782 \pm 0.178$ & $0.670 \pm 0.064$ & $0.715 \pm 0.107$ \\
    2.0 & 1.0 & $0.802 \pm 0.169$ & $0.692 \pm 0.075$ & $0.738 \pm 0.111$ \\
    2.0 & 2.0 & $0.909 \pm 0.092$ & $0.707 \pm 0.031$ & $0.793 \pm 0.045$ \\
    \bottomrule
\end{tabular}
\end{sc}
\end{small}
\end{center}
\vskip -0.1in
\end{table*}

\subsubsection{Other Parameters}\label{sec:method:param:regular}

The chosen values of the remaining parameters are shown in Table~\ref{tab:hparam}. The search for values was in no way exhaustive, focussing mainly on the value of $r_{\mathrm{NMS}}$. An exhaustive search quickly becomes infeasible when considering the number of hyper-parameters and the time required to train the CNN. However, the selected hyper-parameters are sufficiently well-tuned to demonstrate the effectiveness of PPN.

\begin{table}[t]
\caption{Chosen Values of PPN Hyper-Parameters}
\label{tab:hparam}
\begin{center}
\begin{small}
\begin{sc}
\begin{tabular}[c]{rc}
    \toprule
    Parameter           & Value \\
    \midrule
    Input shape         & $(224 \times 224)$ \\
    Origin shape        & $(7 \times 7)$ \\
    $r_{\mathrm{near}}$ & $\sqrt{(2\cdot0.5^2)}$ \\
    $r_{\mathrm{far}}$  & $=r_{\mathrm{near}}$ \\
    $r_{\mathrm{NMS}}$  & 0.35 \\
    $c_{\mathrm{NMS}}$  & 0.8 \\
    \bottomrule
\end{tabular}
\end{sc}
\end{small}
\end{center}
\vskip -0.1in
\end{table}

The input and origin shapes were chosen such that they are compatible with most existing architectures, where $(224 \times 224)$ is a common input shape, and a $(7 \times 7)$ feature map is easy to reach from the input shape. If more dense images are expected, the origin shapes should rather be adjusted, as it will directly affect the number of proposals PPN is able to make. Increasing the input shape may lead to a small increase in speed (as less patches and thus less evaluations are required), but will require more memory to store the model.

$r_{\mathrm{near}}$ was chosen as the minimal radius that is able to cover the entire image, which is the distance from an origin to the centre of a group of four origins. This ensures that no point source is unreachable when constructing the ground truth regression values, while also reducing the number of duplicate proposals.

$r_{\mathrm{far}}$ was minimally experimented with. When $r_{\mathrm{far}} > r_{\mathrm{near}}$, the range in which origins do not contribute to the loss function very clearly affects the results. As the gap increases (i.e. more origins are ignored), the number of false positives increases which drastically reduces precision. Therefore $r_{\mathrm{far}} = r_{\mathrm{near}}$ was chosen, which produced the best results.

The value of $c_{\mathrm{NMS}}$ showed to have very little impact on the results. The specificity of the confidence values is very high, meaning that most guesses provide either a low or a high probability.

\begin{figure}
\begin{center}
    \centerline{\includegraphics[scale=0.5]{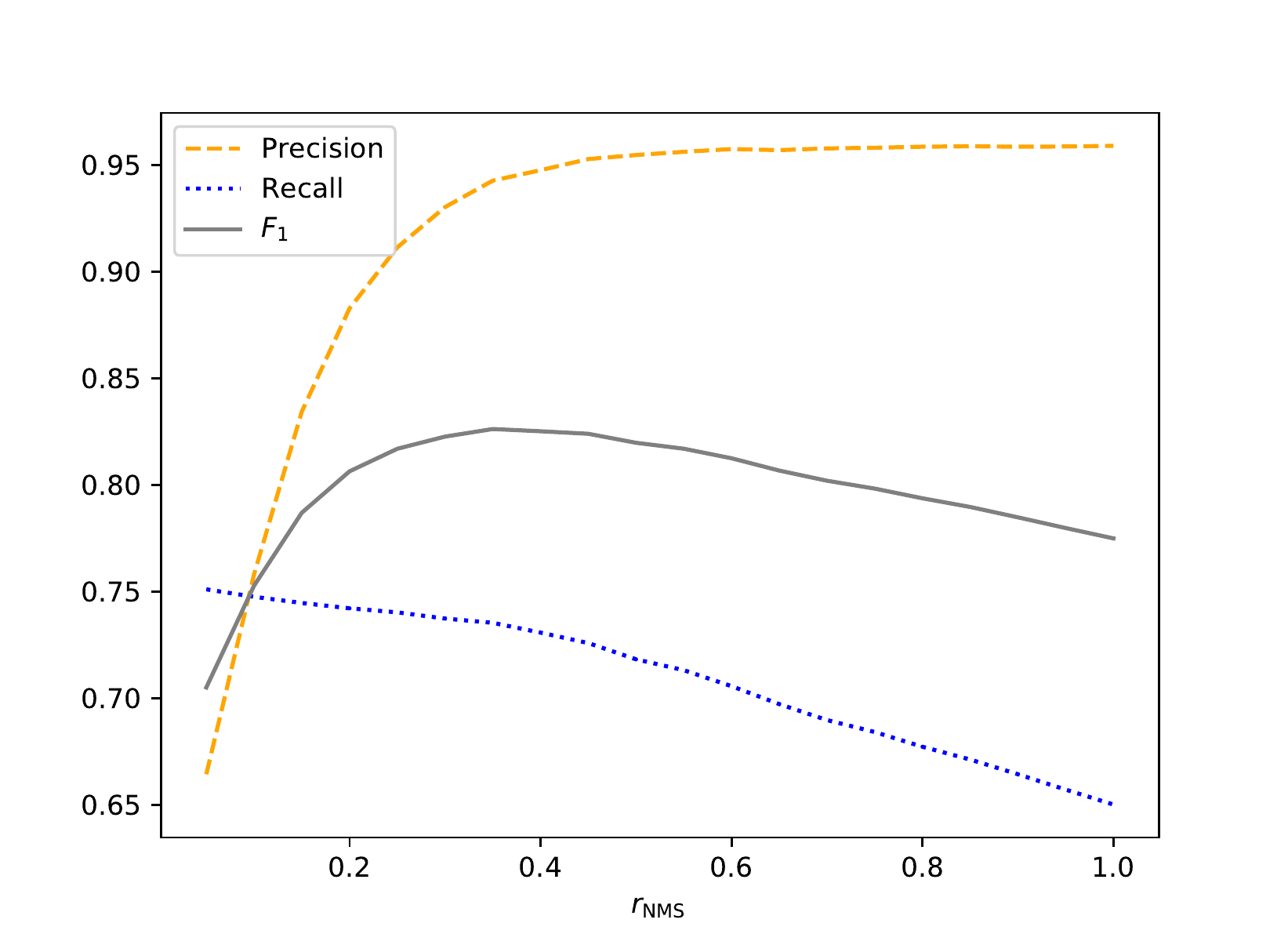}}
    \caption{The precision, recall and $F_1$ scores for PPN when performing NMS with different values of $r_\mathrm{NMS}$. The $F_1$ score is maximised at $r_\mathrm{NMS} = 0.35$.}
    \label{fig:radius-param}
\end{center}
\vskip -0.2in
\end{figure}

Finally, the impact of $r_{\mathrm{NMS}}$ was iteratively explored and shows a strong influence on the balancing of precision and recall. Its effect on precision and recall values is illustrated in Fig.~\ref{fig:radius-param}. As the radius increases (therefore removing more points during NMS), the recall value lowers, but the precision increases. This is a result of removing more duplicates, but also removing some true positives. Conversely, decreasing the radius increases the recall and decreases the precision. Again, this is a result of including more results, and introducing more false positives.

The $r_{\mathrm{NMS}}$ is an interesting parameter to consider when using PPN in practice, seeing the strong influence it has on the precision and recall values. It can therefore be used to help steer the results towards a desired score to maximise. For the purposes of this paper, the $F_1$ score was maximised at $r_{\mathrm{NMS}} = 0.35$.

\section{Experimental Outcomes}\label{sec:results}

This section first describes the method used to generate simulated images in Section~\ref{sec:results:images}. Sections~\ref{sec:results:accuracy} and \ref{sec:results:speed} then respectively compare the accuracy and speed of PPN to that of DeepSource based on different experiments.

\subsection{Image Simulation}\label{sec:results:images}

The true sky can never be known; in order to be used for training, images need to have all sources labelled, including those that humans may have missed. This section provides information on how radio survey images are simulated in order to obtain a sufficiently large, labelled data set.

\begin{table}[t]
\caption{Parameters Used for Simulation of MeerKAT Survey Images}
\label{tab:sim-param}
\begin{center}
\begin{small}
\begin{sc}
\begin{tabular}[c]{lc}
    \toprule
    Parameter                       & Value \\
    \midrule
    Synthesis time                  & 1 h \\
    Integration time                & 5 min \\
    Centre frequency                & 1.4 GHz \\
    Channel width                   & 0.5 MHz \\
    Number of channels              & 10 \\
    System equivalent flux density  & 450 Jy \\
    Pixel scale                     & $5''$ \\
    Clean iterations                & 10000 \\
    Weighting                       & Briggs 1.5 \\
    Number of point sources         & 120-150 \\
    Image size                      & $1024 \mathrm{px} \times 1024 \mathrm{px}$ \\
    \bottomrule
\end{tabular}
\end{sc}
\end{small}
\end{center}
\vskip -0.1in
\end{table}

Images were simulated with the Stimela\footnote{\url{https://github.com/SpheMakh/stimela}} package using the parameters provided in Table~\ref{tab:sim-param}, which is in part similar to those done by Sadr et~al. \cite{Sadr2018}. The parameters simulate observations imaged by the MeerKAT telescope, and have different field centres sampled uniformly at random from $-50^{\circ} < RA < 50^{\circ}, -30^{\circ} < DEC < 50^{\circ}$.

To simulate an image, an empty measurement set which contains the \textit{visibilities} of an interferometer is first created. Visibilities are measurements that represent Fourier components of the sky brightness distribution. Next, it is filled with the expected thermal noise from the MeerKAT setup listed in Table~\ref{tab:sim-param} and then imaged. This creates a ``noise'' image that contains no point sources, from which the background root mean square (rms; equivalent to the standard deviation, $\sigma$), can easily be calculated.

With the background rms calculated, the point sources are generated and added in the image through the Tigger-LSM software\footnote{\url{https://github.com/ska-sa/tigger-lsm}}. Point sources are (uniformly) randomly distributed across the image, allowing sources to overlap. The peak flux, $J$, of the sources is a controlled variable, and is calculated as a multiple of $\sigma$. The peak flux is allocated by separating the sources into 30 equal bins, where the sources in each bin will have an equal flux. The flux for bin $k$ is $J_k = (\frac{1}{3} + \frac{k}{3})\sigma$ for $k \in 0, 1, \ldots ,29$. This provides 30 discrete values for flux in the range of $\frac{1}{3}\sigma$ to $10\sigma$. Examples of generated images are shown in Fig.~\ref{fig:sim-image}.

For use in both PPN and DeepSource, sky coordinates are converted to pixel coordinates to obtain the ground truth locations of all point sources. The flux is linearly scaled across the image such that all values fall in the range $[0, 1]$. For PPN, images are then separated into $(224 \times 224)$ patches with an overlap of 4 pixels.

\begin{table*}
    \centering
    \begin{tabular}{cM{40mm}M{40mm}M{40mm}}
       \toprule
        & $RA=-50^{\circ}, DEC=-30^{\circ}$ & $RA=0^{\circ}, DEC=10^{\circ}$ & $RA=50^{\circ}, DEC=50^{\circ}$ \\
        \midrule
        Image & \includegraphics[width=40mm]{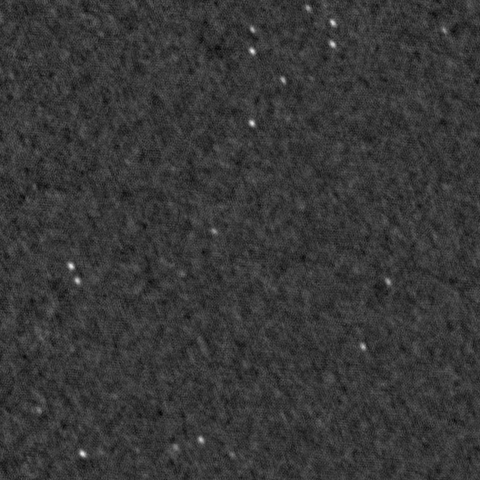} & \includegraphics[width=40mm]{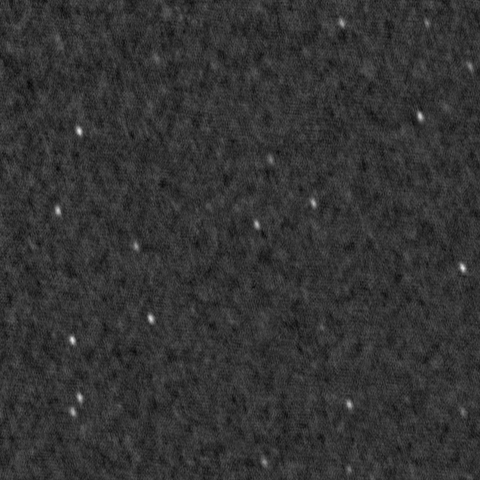} & \includegraphics[width=40mm]{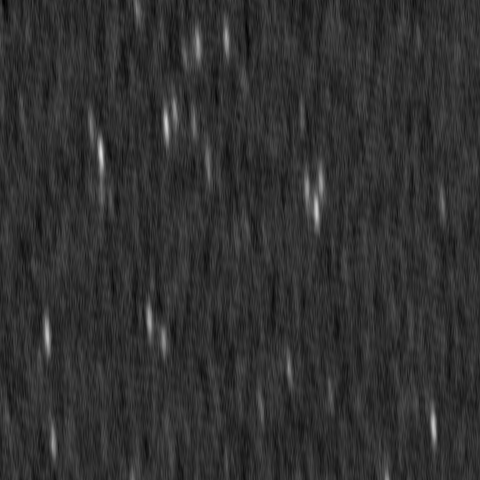} \\
        Ground Truth & \includegraphics[width=40mm]{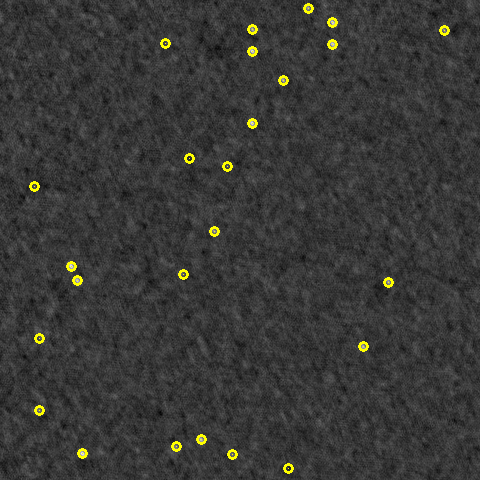} & \includegraphics[width=40mm]{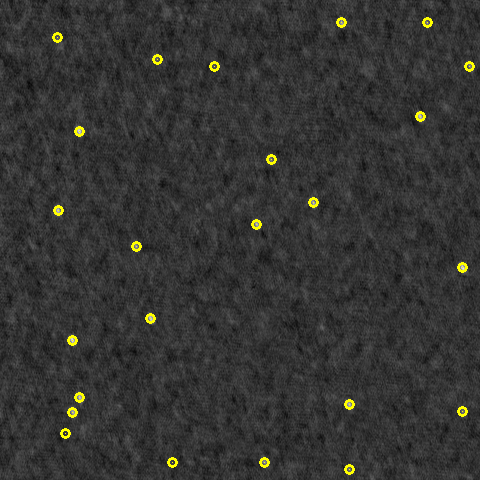} & \includegraphics[width=40mm]{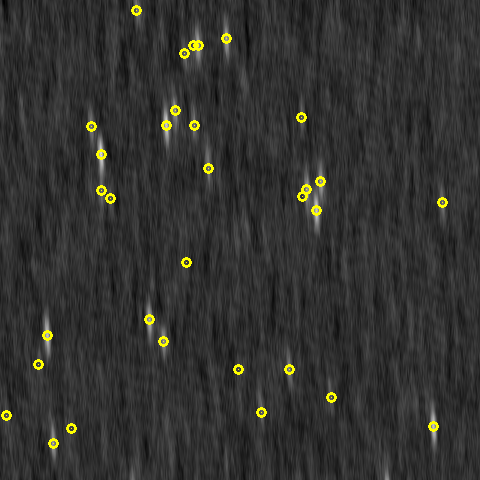} \\
        \bottomrule
    \end{tabular}
    \captionof{figure}{Examples of simulated MeerKAT images. The images were simulated at two corners of the selected field centre range, as well as in the centre of the range. The ground truth positions of point sources are also shown. Note that these images are cropped $(480 \mathrm{px} \times 480\mathrm{px})$ sections from the full $(1024 \mathrm{px} \times 1024 \mathrm{px})$ images.}
    \label{fig:sim-image}
    \vskip -0.1in
\end{table*}

\subsection{Precision and Recall}\label{sec:results:accuracy}

This section compares the precision and recall (as described in Section~\ref{sec:method:metrics}) of PPN to DeepSource. The recall for sources in each bin (i.e. at different flux levels) is calculated individually, which shows the performance of the techniques at different noise levels. The precision is measured across all sources at different values of $r_{tp}$.

For DeepSource, 20 thresholds from 0.1 to 1.0 with logarithmic intervals are used, with $\Delta N_\mathrm{max} = 0$. After identifying islands of pixels, those with a pixel surface area of less than 3 pixels are removed. The centres of the sources are taken as the mean position of the pixels in the island. These hyper-parameters provided maximum recall for sources with $J = 5\sigma$ in the validation set.

\begin{figure*}
    \centering
    \begin{subfigure}[t]{0.48\textwidth}
        \includegraphics[scale=0.5]{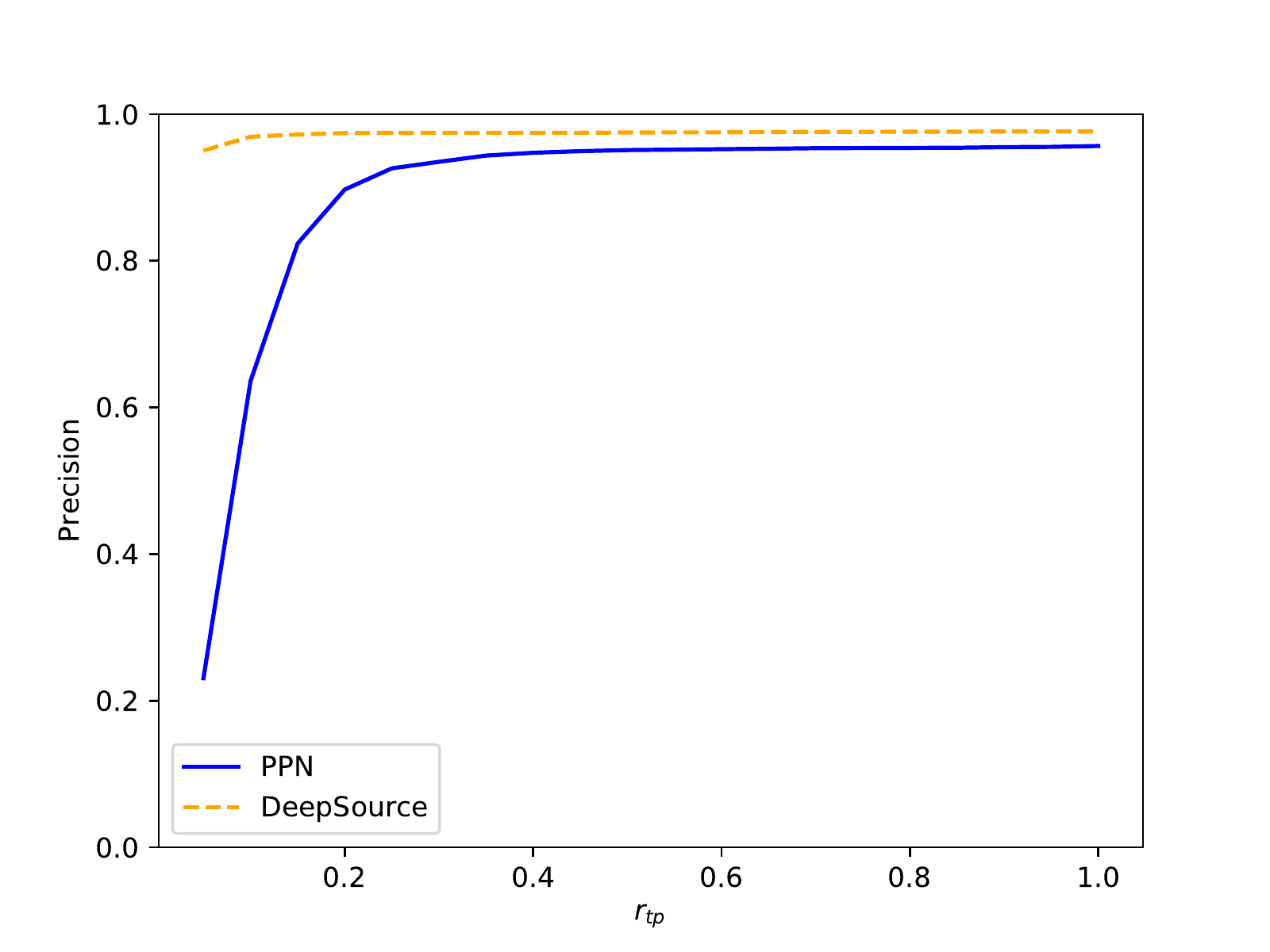}
        \caption{}
        \label{fig:results:prec}
    \end{subfigure}
    \begin{subfigure}[t]{0.48\textwidth}
        \includegraphics[scale=0.5]{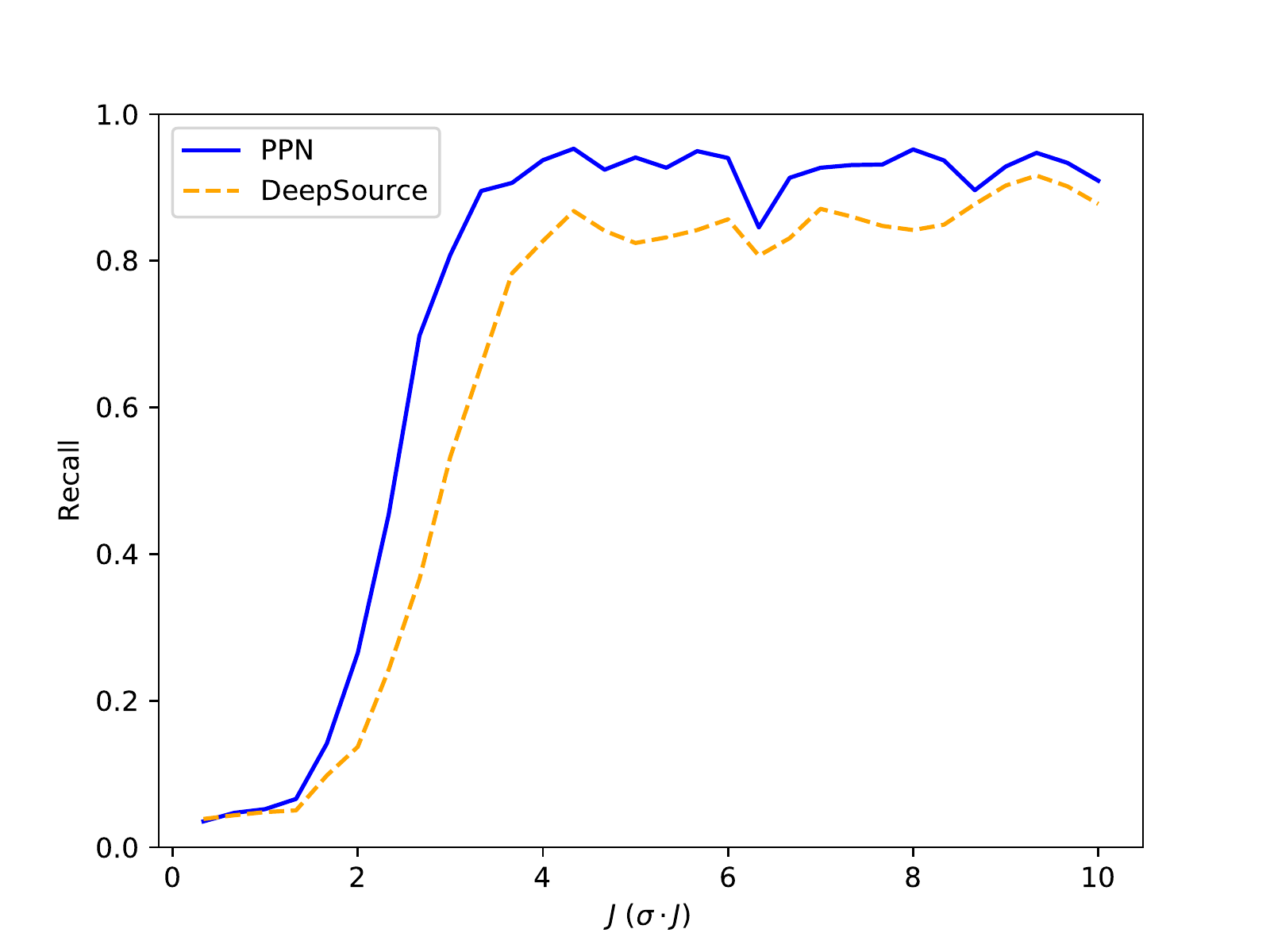}
        \caption{}
        \label{fig:results:recl}
    \end{subfigure}
    \begin{subfigure}[t]{0.48\textwidth}
        \includegraphics[scale=0.5]{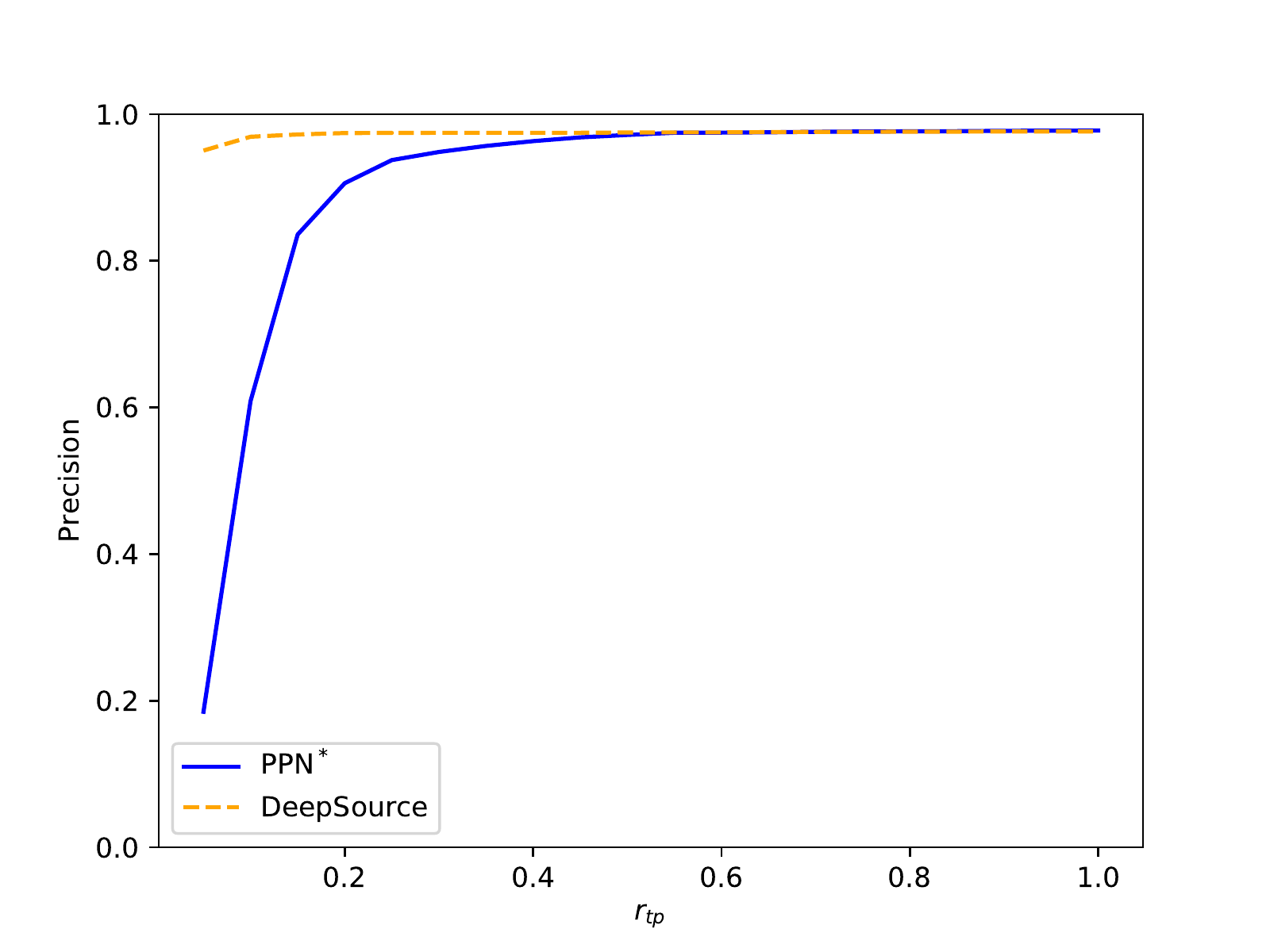}
        \caption{}
        \label{fig:results:prec-star}
    \end{subfigure}
    \begin{subfigure}[t]{0.48\textwidth}
        \includegraphics[scale=0.5]{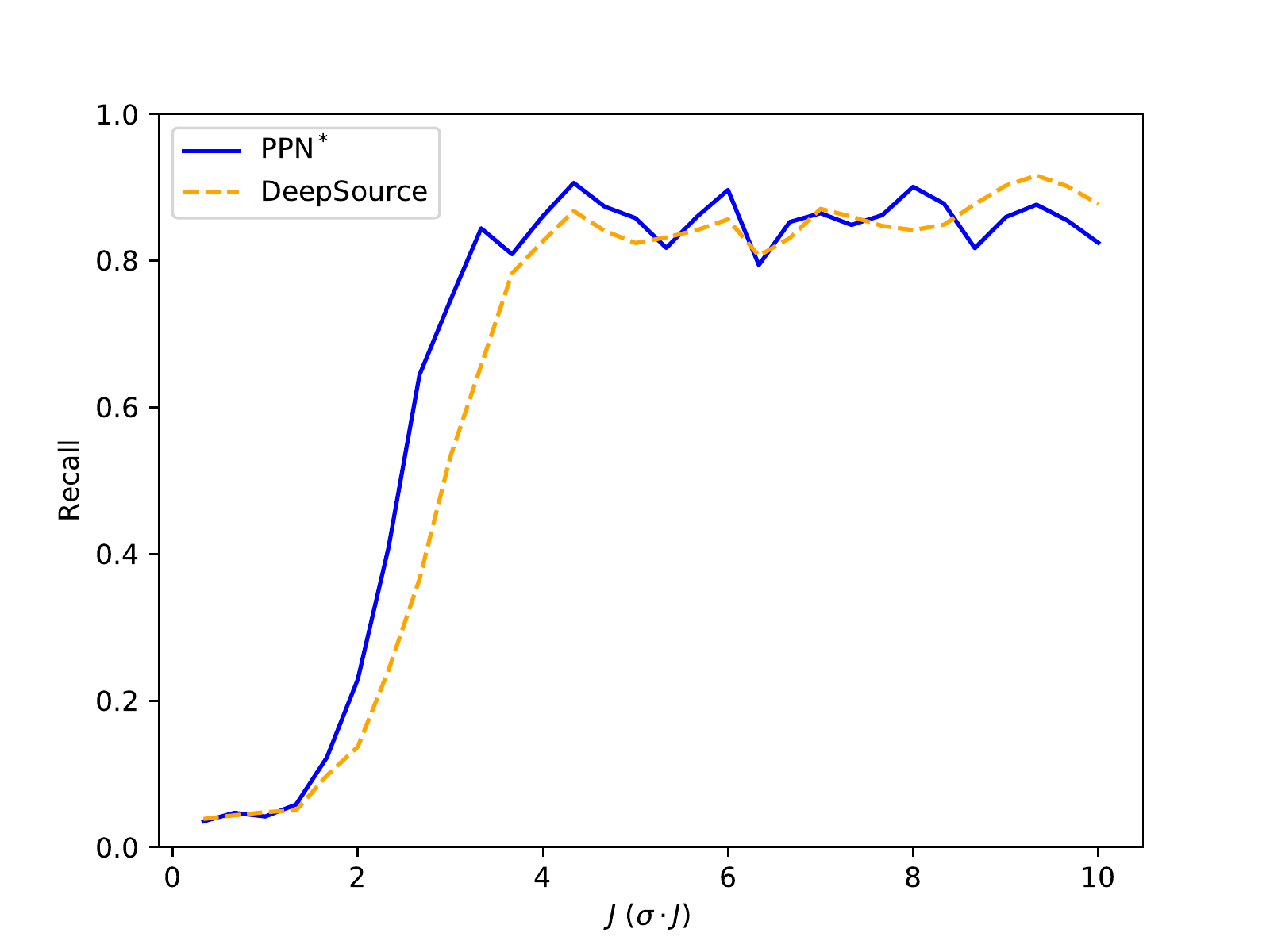}
        \caption{}
        \label{fig:results:recl-star}
    \end{subfigure}
    \caption{Precision and recall scores measured for PPN and DeepSource. Fig.~\ref{fig:results:prec} shows the precision of each method (across all point sources) as a function of $r_{tp}$. This representation of the precision indicates how close to the centres the predictions tend to be, as a smaller value of $r_{tp}$ will only consider closer predictions to be true. Fig.~\ref{fig:results:recl} shows the recall as a function of point source flux. Figs.~\ref{fig:results:prec-star} and~\ref{fig:results:recl-star} show the same measurements for PPN when evaluated with $r_{\mathrm{NMS}} = 0.8$.}
    \label{fig:results}
\end{figure*}

50 new $(1024 \mathrm{px} \times 1024 \mathrm{px})$ images were created for testing the final model. Together, the images contain a total of 6350 sources, providing just over 200 sources in each of the 30 flux level bins. Fig.~\ref{fig:out-image} shows examples of output produced by both PPN and DeepSource. The measured precision and recall scores are shown graphically in Fig.~\ref{fig:results} and listed in Table~\ref{tab:prec-recl}. The listing contains only partial results for the sake of brevity.

Fig.~\ref{fig:results:recl} shows that both PPN and DeepSource perform well at levels with $J \ge 5\sigma$, with the recall worsening at $4\sigma$ and performing poorly at lower levels. Both methods are able to recall more than 80\% of sources at $J \ge 4\sigma$, however PPN maintains a higher recall at all levels except the very lowest ($J = 0.3\sigma$).

The precision in Fig.~\ref{fig:results:prec} shows excellent results for both methods, although DeepSource dominates PPN in this regard. The graph shows that DeepSource in general has a higher precision than PPN, and also detects the source centres closer to the ground truth centre. This is indicated by the slower decline at lower values of $r_{tp}$.

One might wonder how larger values of $r_{\mathrm{NMS}}$ (explored in Section~\ref{sec:method:param:regular}) affects both the recall and precision on the final outcomes. To this end, PPN was additionally evaluated with $r_{\mathrm{NMS}} = 0.8$, denoted as PPN$^*$ in both Fig.~\ref{fig:results} and Table~\ref{tab:prec-recl}. The results show a higher precision and a recall closer to what DeepSource achieves. The precision of course still drops at lower values since the regression is no more accurate than before, but the precision goes up as more false positives are removed.

These experiments show that PPN is capable of proposing point sources fairly accurately. The recall is comparable to that of DeepSource, however the precision is worse due to the inaccuracy of regression. When reliability is the most important aspect, PPN should not be the first choice, however the next experiment shows the trade-off that could be made in favour of speed which will be necessary for SKA pathfinders as the survey image sizes increase.

\begin{table*}
    \centering
    \begin{tabular}{cM{33mm}M{33mm}M{33mm}M{33mm}}
       \toprule
        & Original & Ground Truth & DeepSource & PPN \\
        \midrule
        (a) & \includegraphics[width=33mm]{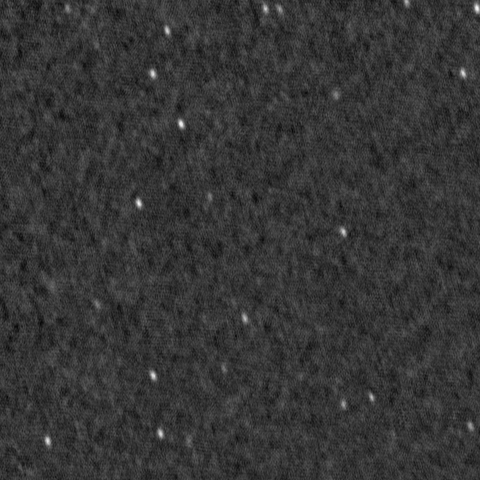} & \includegraphics[width=33mm]{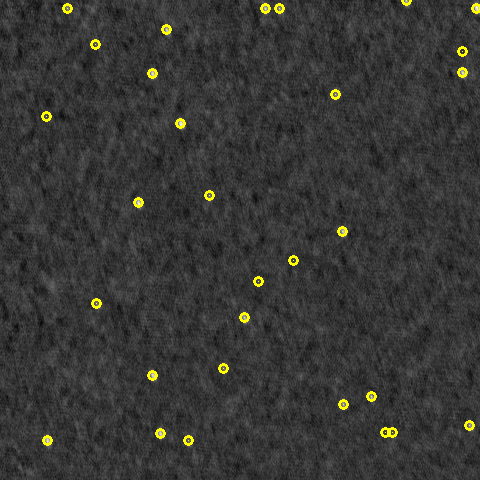} & \includegraphics[width=33mm]{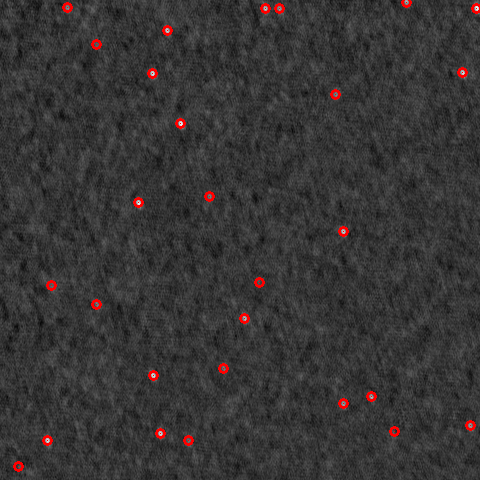} & \includegraphics[width=33mm]{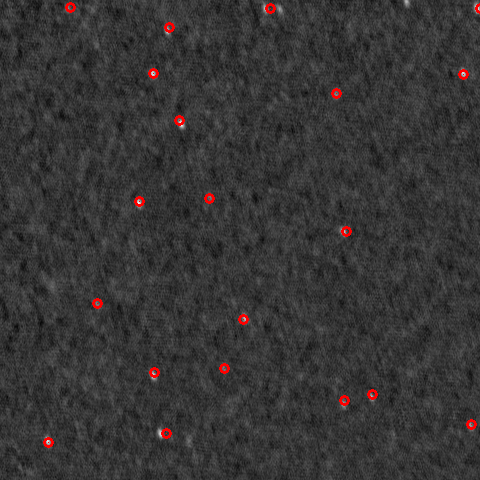} \\
        (b) & \includegraphics[width=33mm]{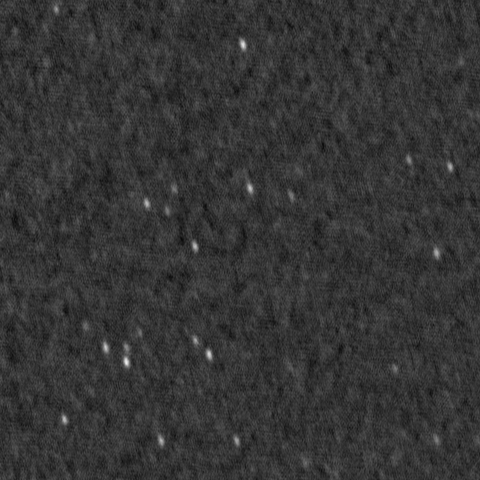} & \includegraphics[width=33mm]{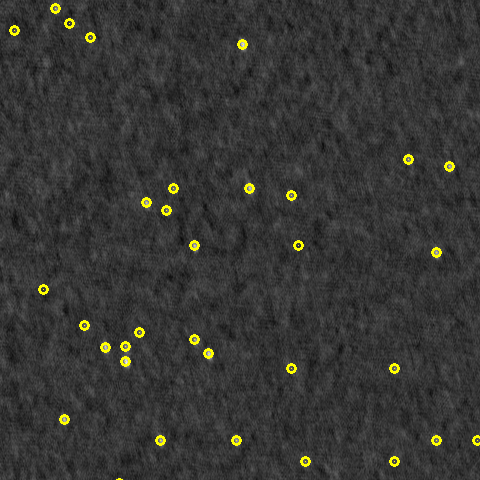} & \includegraphics[width=33mm]{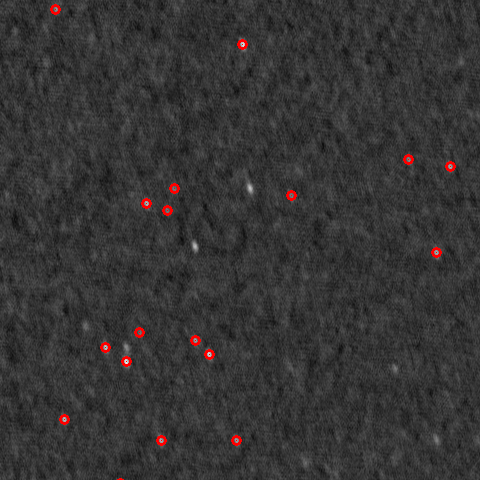} & \includegraphics[width=33mm]{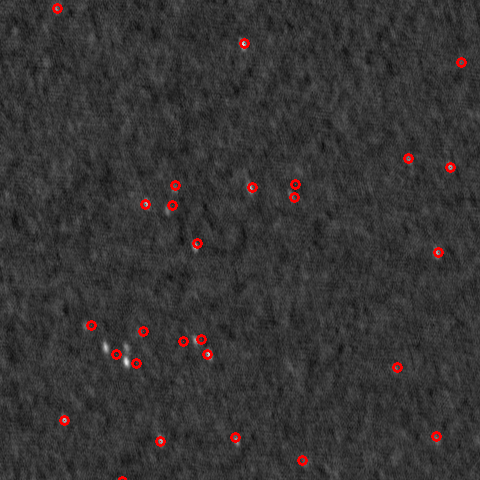} \\
        (c) & \includegraphics[width=33mm]{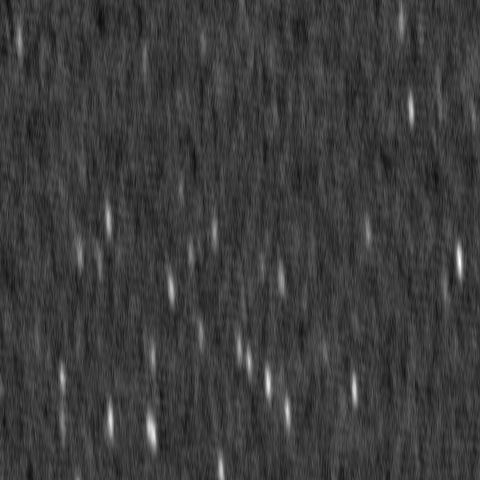} & \includegraphics[width=33mm]{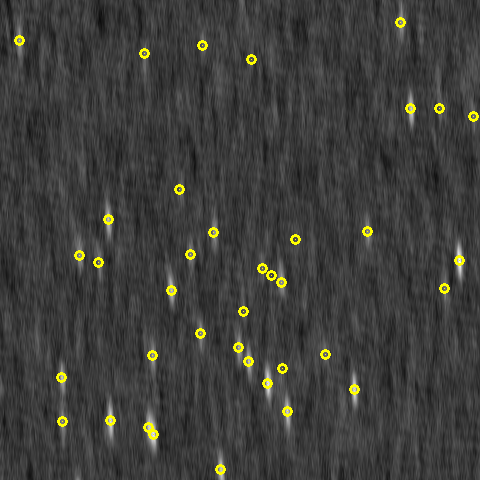} & \includegraphics[width=33mm]{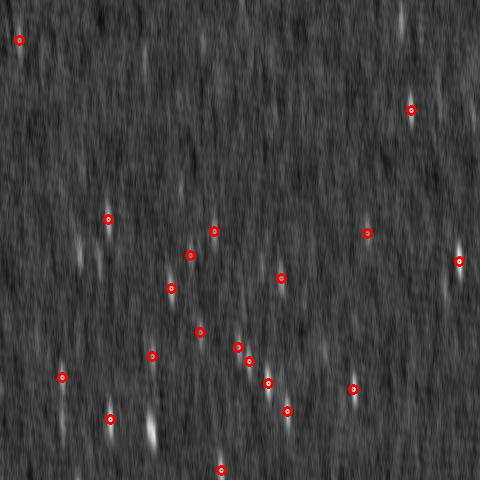} & \includegraphics[width=33mm]{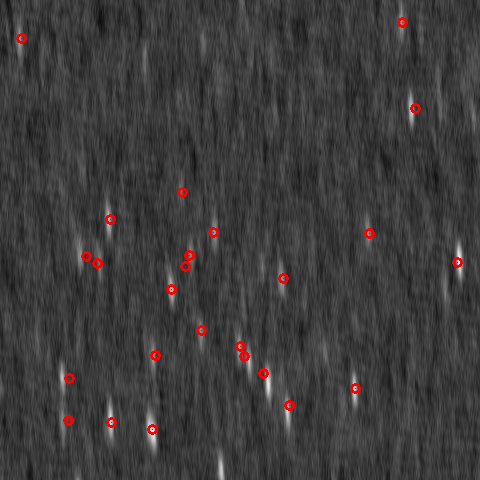} \\
        \bottomrule
    \end{tabular}
    \captionof{figure}{Example outputs produced by DeepSource and PPN. The precision of DeepSource is evident in these examples as the detected point centres are almost always right on the true centre. PPN is in most cases at least on the point source, but is further away from the true centres. A few shortcomings that can be seen here is that DeepSource tends to miss more sources which leads to its lower recall, and PPN suffers from bad precision in dense clusters, as can be seen in image (b). Note that these images are cropped $(480 \times 480)$ sections from the full $(1024 \times 1024)$ images.}
    \label{fig:out-image}
    \vskip -0.1in
\end{table*}

\begin{table}
\vskip 0.2in
\caption{Precision and Recall Scores Measured for DeepSource and PPN}
\label{tab:prec-recl}
\begin{center}
\begin{small}
\begin{sc}
\begin{tabular}[c]{lccc}
    \toprule
    \multicolumn{4}{c}{Precision} \\
    \midrule
    $r_{tp}$ & DeepSource & PPN & PPN$^*$ \\
    \midrule
    $0.05$ & $\mathbf{0.950 \pm 0.083}$ & $0.231 \pm 0.048$ & $0.185 \pm 0.043$ \\
    $0.10$ & $\mathbf{0.969 \pm 0.068}$ & $0.636 \pm 0.066$ & $0.609 \pm 0.073$ \\
    $0.15$ & $\mathbf{0.972 \pm 0.067}$ & $0.824 \pm 0.062$ & $0.836 \pm 0.062$ \\
    $0.20$ & $\mathbf{0.974 \pm 0.064}$ & $0.897 \pm 0.052$ & $0.906 \pm 0.048$ \\
    $0.25$ & $\mathbf{0.975 \pm 0.064}$ & $0.926 \pm 0.049$ & $0.937 \pm 0.046$ \\
    $0.40$ & $\mathbf{0.975 \pm 0.063}$ & $0.947 \pm 0.040$ & $0.963 \pm 0.038$ \\
    $0.50$ & $\mathbf{0.975 \pm 0.063}$ & $0.951 \pm 0.041$ & $0.972 \pm 0.037$ \\
    $1.00$ & $0.976 \pm 0.060$ & $0.957 \pm 0.039$ & $\mathbf{0.978 \pm 0.032}$ \\
    \toprule
    \multicolumn{4}{c}{Recall} \\
    \midrule
    $J\ (\cdot \sigma)$ & DeepSource & PPN & PPN$^*$ \\
    \midrule
    $0.3$ & $\mathbf{0.039 \pm 0.092}$ & $0.035 \pm 0.090$ & $0.035 \pm 0.090$ \\
    $0.7$ & $0.044 \pm 0.096$ & $\mathbf{0.047 \pm 0.092}$ & $\mathbf{0.047 \pm 0.092}$ \\
    $1.0$ & $0.048 \pm 0.106$ & $\mathbf{0.052 \pm 0.101}$ & $0.042 \pm 0.085$ \\
    $1.3$ & $0.050 \pm 0.114$ & $\mathbf{0.066 \pm 0.115}$ & $0.058 \pm 0.115$ \\
    $1.7$ & $0.098 \pm 0.182$ & $\mathbf{0.142 \pm 0.204}$ & $0.123 \pm 0.157$ \\
    $2.0$ & $0.137 \pm 0.231$ & $\mathbf{0.265 \pm 0.234}$ & $0.229 \pm 0.204$ \\
    $2.3$ & $0.242 \pm 0.324$ & $\mathbf{0.453 \pm 0.297}$ & $0.409 \pm 0.266$ \\
    $2.7$ & $0.366 \pm 0.387$ & $\mathbf{0.698 \pm 0.299}$ & $0.644 \pm 0.303$ \\
    $3.0$ & $0.533 \pm 0.362$ & $\mathbf{0.808 \pm 0.229}$ & $0.746 \pm 0.250$ \\
    $4.0$ & $0.827 \pm 0.242$ & $\mathbf{0.937 \pm 0.130}$ & $0.861 \pm 0.197$ \\
    $5.0$ & $0.824 \pm 0.266$ & $\mathbf{0.941 \pm 0.121}$ & $0.858 \pm 0.168$ \\
    $6.0$ & $0.857 \pm 0.232$ & $\mathbf{0.940 \pm 0.121}$ & $0.897 \pm 0.155$ \\
    $7.0$ & $0.871 \pm 0.216$ & $\mathbf{0.927 \pm 0.116}$ & $0.865 \pm 0.156$ \\
    $8.0$ & $0.842 \pm 0.215$ & $\mathbf{0.952 \pm 0.102}$ & $0.901 \pm 0.146$ \\
    $9.0$ & $0.903 \pm 0.220$ & $\mathbf{0.929 \pm 0.133}$ & $0.860 \pm 0.167$ \\
    $10.0$ & $0.878 \pm 0.198$ & $\mathbf{0.909 \pm 0.143}$ & $0.825 \pm 0.199$ \\
    \bottomrule
\end{tabular}
\end{sc}
\end{small}
\end{center}
\vskip 0.5in 
\end{table}

\subsection{Evaluation Speed}\label{sec:results:speed}

The poor scaling ability of point source detectors is a large motivation for the development of PPN. This section breaks down the inference time (i.e. compute time) of both PPN and DeepSource\footnote{All experiments were performed on the Ilifu infrastructure using a single Nvidia Tesla P100 12GB.}. All time measurements are in wall-clock seconds.

For the experiment, images from size $(1024 \mathrm{px} \times 1024 \mathrm{px})$ up to $(16384 \mathrm{px} \times 16384 \mathrm{px})$ were generated, with each category being 1024 pixels larger on both dimensions than the previous. This results in 16 size categories, for which 50 images were generated each. The number of point sources are also increased as to maintain a similar density of sources across the image surface.

Timing starts when the model has been created and the image already resides in memory. The times recorded therefore exclude the time of creating the model as it need only be done once, and excludes the time taken to load the image into main memory as it affects both techniques equally. Times do however include the copying of data to and from the GPU.

A modification of DeepSource was necessary to cope with memory constraints. DeepSource is defined to handle one image at a time, however the CNN model grows drastically as the images become larger. For this reason, the model was limited to dealing with $(4096 \mathrm{px} \times 4096 \mathrm{px})$ images. For larger images, the image was patched in the same manner as for PPN. The patches are individually fed through the CNN and the results are integrated back with overlapping sections being averaged.

The individual steps of each technique are recorded separately to provide a detailed breakdown of where computation is spent. For PPN, these include the patching of the original image, the evaluation of the CNN, and non-maximum suppression for removal of duplicates. For DeepSource, these include the evaluation of the CNN, thresholded blob detection (TBD), and also additional patching and reintegration of the image as explained above.

It is difficult to measure a precise time for TBD, as the number of iterations may vary between images based on $\Delta N_{max}$ and the number of thresholds used. Additionally, TBD iterations may be parallelised, providing some speed-up. For this reason, only a single iteration of TBD is performed at $\tau = 0.5$. This decision can be seen as measuring a lower bound estimate of the DeepSource implementation's computational time.

\begin{figure}
\vskip -0.4in
\begin{center}
    \centerline{\includegraphics[scale=0.5]{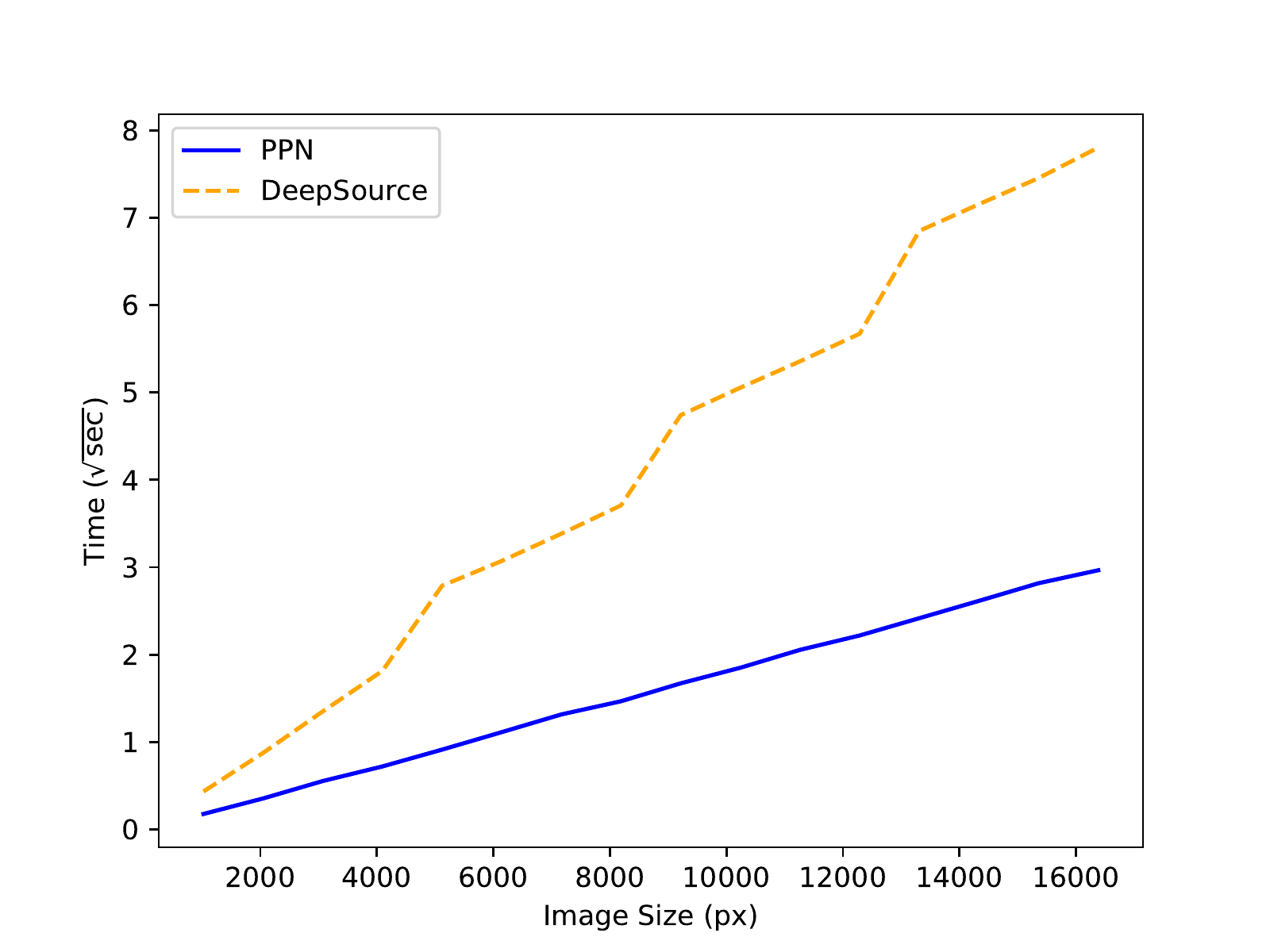}}
    \caption{The measured total compute times for PPN and DeepSource at different image sizes. The figure takes the square root scaled time against the image size, which gives an indication of the per-pixel time complexity.}
    \label{fig:bench}
\end{center}
\vskip -0.2in
\end{figure}

The measured total inference times are illustrated in Fig.~\ref{fig:bench}. The times measured for selected image sizes and the individual steps are listed in Table~\ref{tab:bench}. The listing only contains select results for the sake of brevity. From the results, PPN shows better speed at both smaller and larger images when compared to DeepSource. The largest tested images at $(16384 \mathrm{px} \times 16384 \mathrm{px})$ are processed by PPN in under 20\% of the time used by DeepSource.

Fig.~\ref{fig:bench} suggests that PPN has a time complexity of $\mathcal{O}(ij)$ for an image of size $(i \times j)$, though more detailed analysis will be necessary to confirm. DeepSource has a slightly worse complexity as a result of the flood-fill algorithm. The step pattern present in the DeepSource timings indicates the effect of the model's size limit. Every time an image becomes sufficiently large, more individual evaluations of the CNN are required, leading to an increase in the total CNN evaluation time.

\begin{figure*}
\begin{center}
    {\begin{subfigure}[t]{0.48\textwidth}
        \includegraphics[scale=0.5]{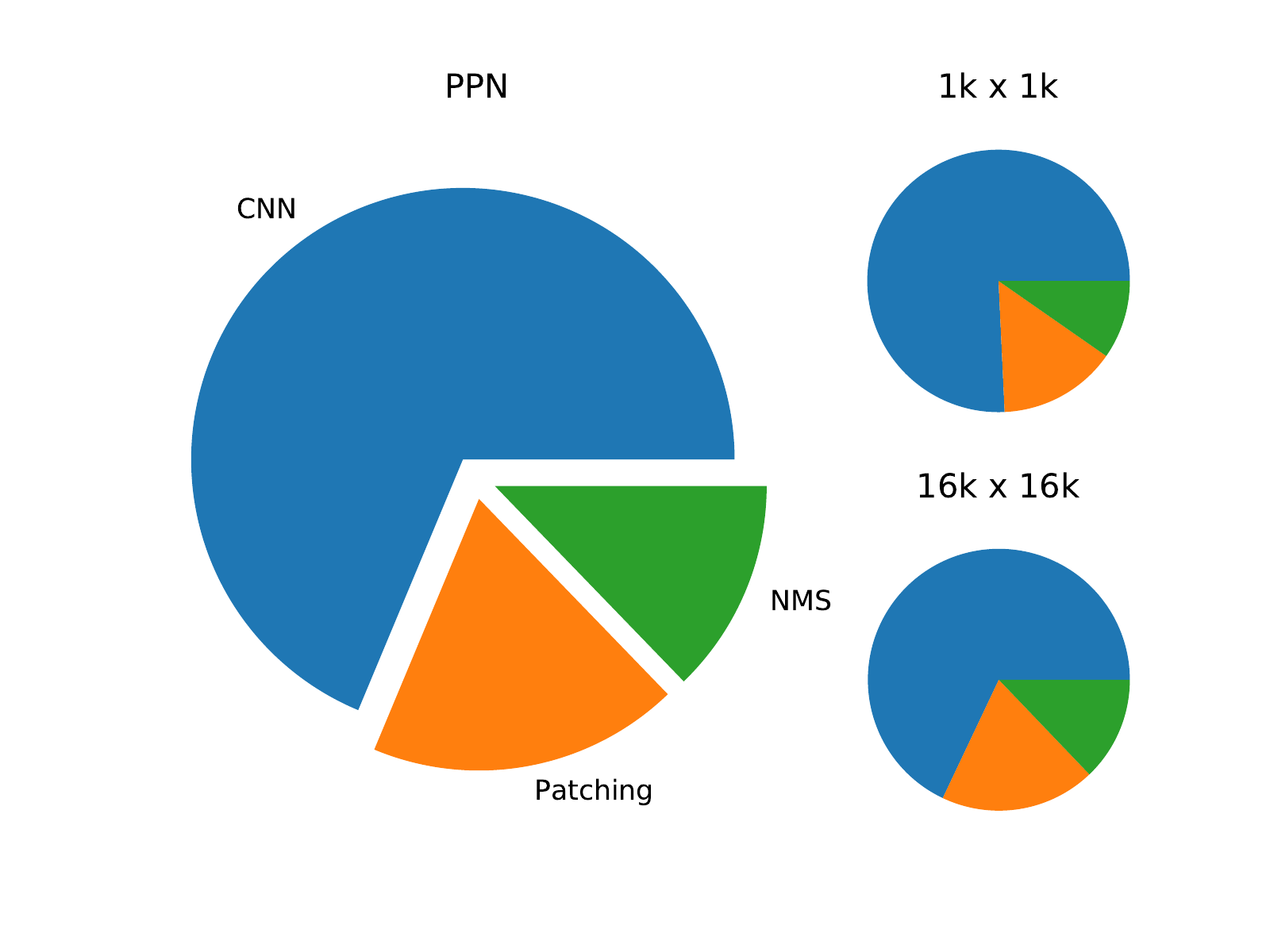}
        \caption{}\label{fig:pie:ppn}
    \end{subfigure}
    \begin{subfigure}[t]{0.48\textwidth}
        \includegraphics[scale=0.5]{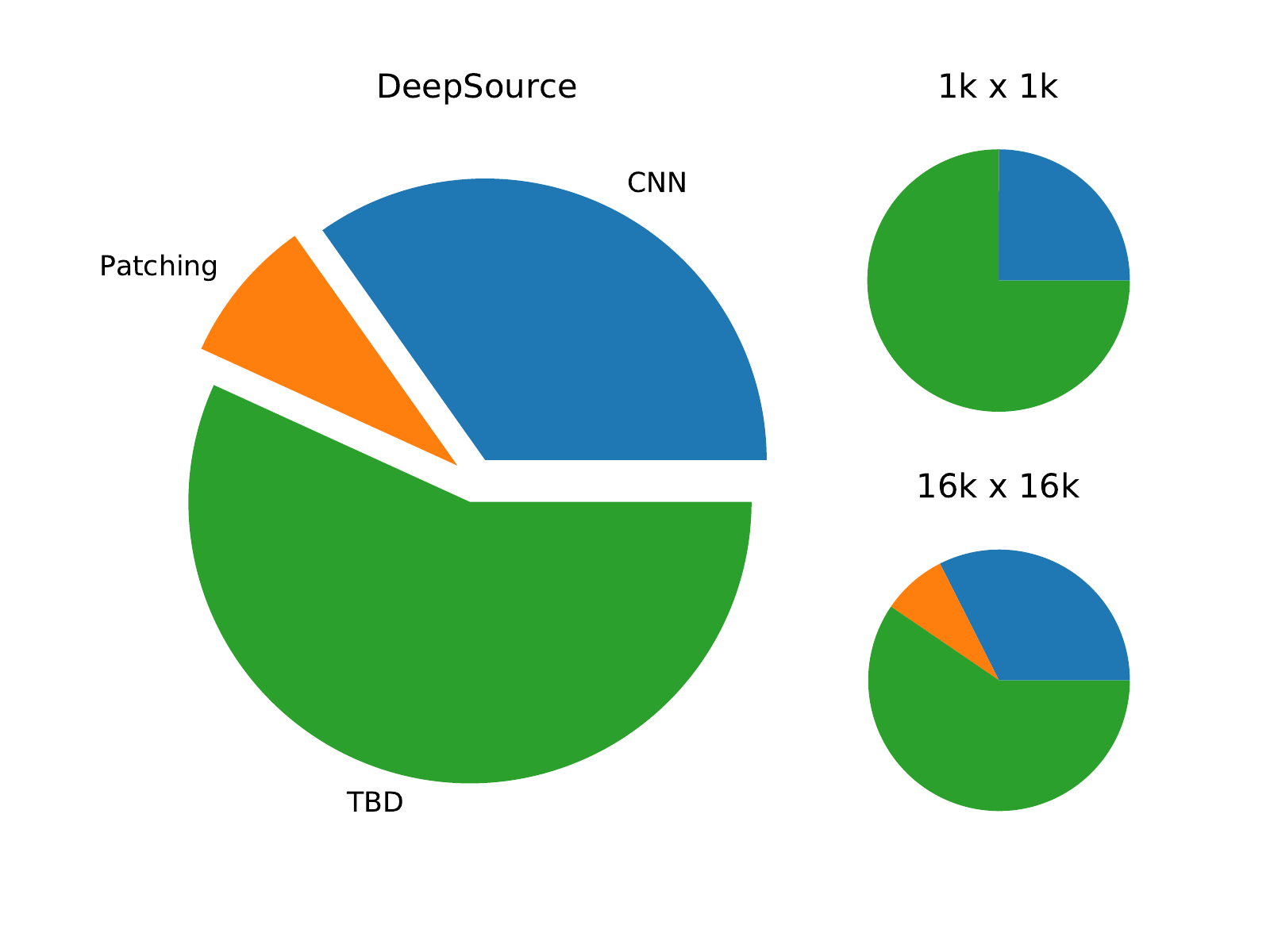}
        \caption{}\label{fig:pie:ds}
    \end{subfigure}}
    \caption{A visualisation of the mean compute time compositions for PPN and DeepSource. Fig.~\ref{fig:pie:ppn} shows the composition for PPN and \ref{fig:pie:ds} for DeepSource. The compositions for the smallest and largest images are also shown in each case.}
    \label{fig:pie}
\end{center}
\vskip -0.2in
\end{figure*}

Fig.~\ref{fig:pie:ppn} shows a breakdown of the PPN steps. The CNN evaluation takes up the majority of computation time, as can be expected due to the depth of the CNN and the simplicity of the other steps involved. However, as the images become larger, the fraction of time being spent on the patching and NMS becomes larger, though the majority of time is still spent on CNN evaluation.

Fig.~\ref{fig:pie:ds} shows a breakdown of the DeepSource steps. Here the CNN evaluation time is less than the single TBD iteration. This emphasises the poor complexity of the flood fill algorithm, the effects of which will worsen with more TBD iterations. The time spent patching the image is longer than the patching performed by PPN, due to the reintegration step after CNN evaluation.

PPN is able to process images faster than the current state of the art, machine learning based approach. The performance gain also only becomes more prominent as the images become larger, providing a more scalable approach.

\begin{table*}
\caption{Measured Evaluation Times (sec) for PPN and DeepSource at Different Image Sizes}
\label{tab:bench}
\begin{center}
\begin{small}
\begin{sc}
\begin{tabular}[c]{lcccc}
    \toprule
    \multicolumn{5}{c}{PPN} \\
    \midrule
    Size & Patching & CNN & NMS & Total \\
    \midrule
    $1024^2$ & $0.004 \pm 0.000$ & $0.032 \pm 0.038$ & $0.005 \pm 0.037$ & $0.041 \pm 0.002$ \\
    $2048^2$ & $0.018 \pm 0.003$ & $0.095 \pm 0.021$ & $0.012 \pm 0.018$ & $0.125 \pm 0.011$ \\
    $4096^2$ & $0.075 \pm 0.004$ & $0.381 \pm 0.035$ & $0.063 \pm 0.028$ & $0.519 \pm 0.013$ \\
    $8192^2$ & $0.374 \pm 0.022$ & $1.500 \pm 0.023$ & $0.277 \pm 0.035$ & $2.151 \pm 0.036$ \\
    $12288^2$ & $0.925 \pm 0.031$ & $3.362 \pm 0.037$ & $0.636 \pm 0.050$ & $4.924 \pm 0.063$ \\
    $16384^2$ & $1.687 \pm 0.032$ & $5.976 \pm 0.068$ & $1.130 \pm 0.086$ & $8.793 \pm 0.097$ \\
    \toprule
    \multicolumn{5}{c}{DeepSource} \\
    \midrule
    Size & Patching & CNN & TBD & Total \\
    \midrule
    $1024^2$ & N/A & $0.045 \pm 0.002$ & $0.137 \pm 0.002$ & $0.188 \pm 0.004$ \\
    $2048^2$ & N/A & $0.177 \pm 0.009$ & $0.555 \pm 0.009$ & $0.767 \pm 0.014$ \\
    $4096^2$ & N/A & $0.776 \pm 0.005$ & $2.281 \pm 0.141$ & $3.290 \pm 0.141$ \\
    $8192^2$ & $1.245 \pm 0.012$ & $3.476 \pm 0.017$ & $9.026 \pm 0.129$ & $13.747 \pm 0.133$ \\
    $12288^2$ & $2.748 \pm 0.023$ & $9.089 \pm 0.055$ & $20.329 \pm 0.284$ & $32.166 \pm 0.282$ \\
    $16384^2$ & $4.884 \pm 0.114$ & $19.751 \pm 2.778$ & $36.233 \pm 0.938$ & $60.868 \pm 3.000$ \\
    \bottomrule
\end{tabular}
\end{sc}
\end{small}
\end{center}
\vskip -0.1in
\end{table*}

\section{Conclusion}\label{sec:conclusion}

This paper proposed the Point Proposal Network, which is a point source detection technique that uses modern object-detection techniques based on deep CNNs. The technique addresses the slow processing speeds and poor scalability of existing techniques.

PPN was compared to DeepSource, the state of the art machine learning based approach to point source detection. Experiments showed that PPN has a similar ability to recall point sources, though the precision is weaker. Further experiments showed that PPN is able to process images much faster than DeepSource and is able to scale to larger images.

There is a lot of room to further develop and investigate the PPN approach. PPN can possibly be used in a larger pipeline, where PPN is first used to estimate where point sources are. The estimates could then be used to produce sub-images on which other methods can be used to produce more reliable results. PPN can also be expanded by including additional characteristics in the regression output. This may include the variances of a Gaussian distribution fitting the source, or even possible classification scores. Many different possibilities can be explored.

The significant source detection challenges that will be faced by the SKA and its pathfinders require a suite of robust, sophisticated and scalable methods to fully extract the scientific value of increasingly deep views of the Universe. PPN could well contribute to that with further testing and development.

\ifCLASSOPTIONcompsoc
  \section*{Acknowledgments}
\else
  \section*{Acknowledgment}
\fi

The authors would like to thank the Inter-University Institute for Data-Intensive Astronomy (IDIA) for sponsoring the research and providing the computational infrastructure used throughout.


\newpage

\bibliographystyle{IEEEtran}
\bibliography{bib}


\begin{IEEEbiography}[{\includegraphics[width=1in,height=1.25in,clip,keepaspectratio]{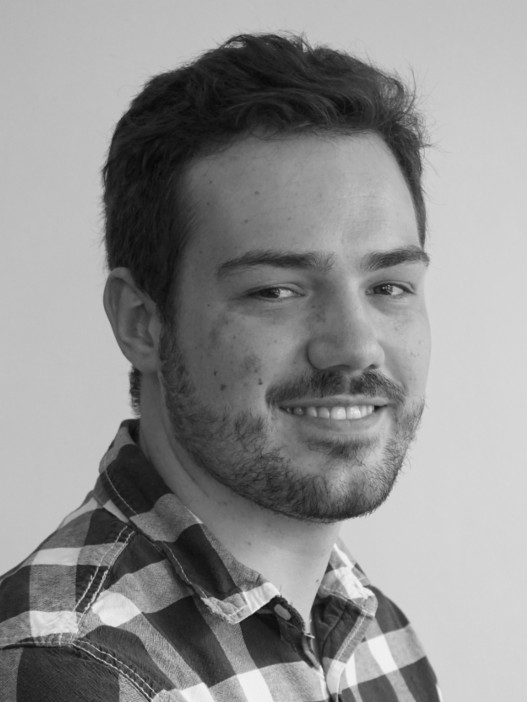}}]{Duncan Tilley}
received his Bachelor's Degree (Hons) in Computer Science from the University of Pretoria in 2019. While currently working as a Software Engineer, he is continuing his research at the University of Pretoria in pursuit of a Master's degree. His research interests include machine learning, radio astronomy and swarm intelligence.
\end{IEEEbiography}

\begin{IEEEbiography}[{\includegraphics[width=1in,height=1.25in,clip,keepaspectratio]{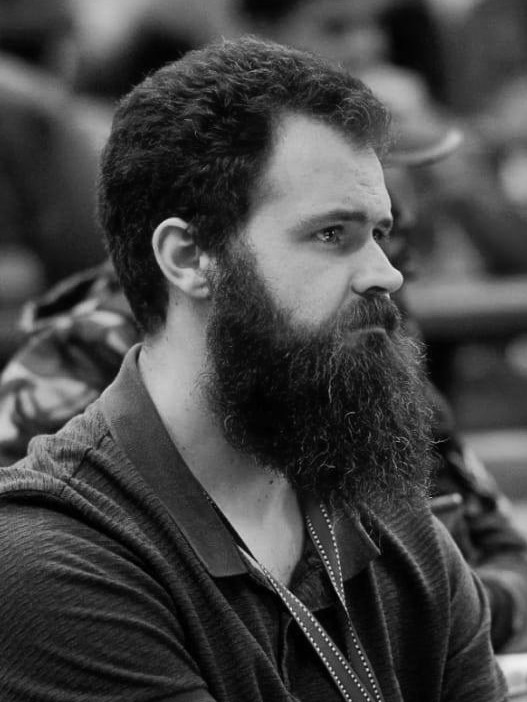}}]{Christopher W. Cleghorn}
received his Master's and PhD degrees in Computer Science from the University of Pretoria, South Africa, in 2013 and 2017 respectively. He is a Senior lecturer in Computer Science at the University of Pretoria. His research interests include swarm intelligence, evolutionary computation, machine learning, and radio astronomy with a strong focus of theoretical research. Dr Cleghorn annually serves as a reviewer for numerous international journals and conferences in domains ranging from swarm intelligence and neural networks to mathematical optimization.
\end{IEEEbiography}

\begin{IEEEbiography}[{\includegraphics[width=1in,height=1.25in,clip,keepaspectratio]{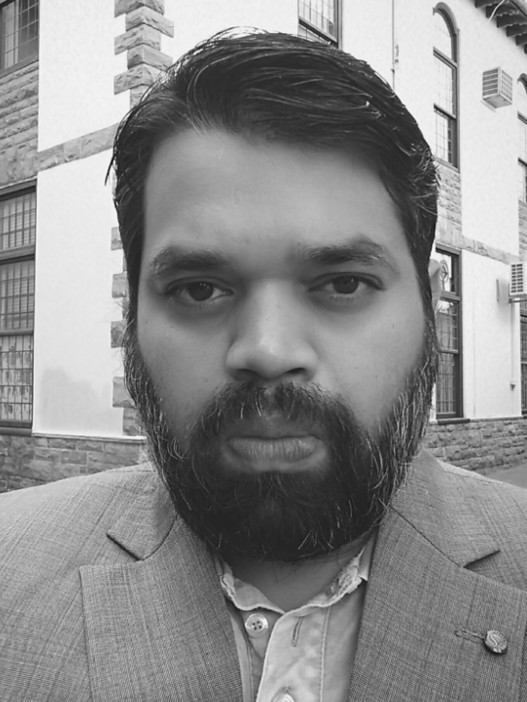}}]{Kshitij Thorat}
is a radio astronomer and has worked as a postdoctoral fellow at the University of Pretoria since 2019. He finished his doctoral work at the Indian Institute of Science in 2014, after which he has been based in South Africa. His research interests revolve around  extra-galactic radio sources: their life-cycles, morphology and their impact on their surroundings. With the advent of Big Data paradigm in radio astronomy, Kshitij has focused on techniques to help process and analyse data better, including development of fully automated data reduction pipelines and machine learning techniques to classify radio galaxies.
\end{IEEEbiography}

\begin{IEEEbiography}[{\includegraphics[width=1in,height=1.25in,clip,keepaspectratio]{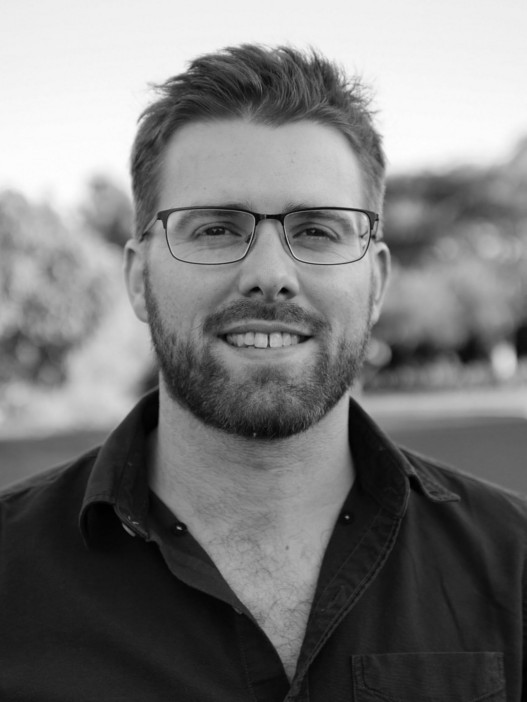}}]{Roger Deane}
is an associate professor at the University of Pretoria. He completed his doctorate in 2012 at the University of Oxford. In 2018 he moved to the University of Pretoria where he established the astronomy research group. His research interests cover a broad range of energy and spatial scales, from diffuse neutral hydrogen in galaxies to the inner accretion disks of supermassive black holes, using the power of next-generation radio telescopes such as South Africa’s MeerKAT radio telescope, the precursor to the Square Kilometre Array, and the Event Horizon Telescope.
\end{IEEEbiography}

\end{document}